\documentclass[10pt,twocolumn,letterpaper]{article}

\usepackage[pagenumbers]{cvpr} % To force page numbers, e.g. for an arXiv version
\usepackage{multirow}
\usepackage[table,dvipsnames]{xcolor}
\usepackage{booktabs}
\usepackage{tabularx} % also loads 'array' package
\usepackage{tabulary}
\definecolor{darkgreen}{rgb}{0.54, 0.76, 0.29}
\definecolor{lightgreen}{rgb}{0.77, 0.88, 0.65}

\newcommand\blfootnote[1]{%
\begingroup
\renewcommand\thefootnote{}\footnote{#1}%
\addtocounter{footnote}{-1}%
\endgroup
}

\usepackage{tabularx}
\usepackage{float}

\newcommand{\method}{{Seq2Time}\xspace}
\usepackage{pifont}

\definecolor{customblue}{HTML}{357ABD} % Replace with your color's hex code

\newcommand{\cmark}{\ding{51}}
\newcommand{\xmark}{\ding{55}}
\newcommand{\lgcmark}{\textcolor{green}{\cmark}}
\newcommand{\lgxmark}{\textcolor{red}{\xmark}}

\definecolor{cvprblue}{rgb}{0.21,0.49,0.74}
\usepackage[pagebackref,breaklinks,colorlinks,allcolors=cvprblue]{hyperref}
\usepackage{amsmath}
\usepackage{amsfonts}
\usepackage{amssymb}
\usepackage{graphicx}

\title{\method: Sequential Knowledge Transfer for Video LLM \\ Temporal Grounding}
\author{
    Andong Deng$^{1,2, *}$, Zhongpai Gao$^{2,\dagger}$, Anwesa Choudhuri$^{2}$, Benjamin Planche$^{2}$, Meng Zheng$^{2}$, \\
    Bin Wang$^{2,3}$, Terrence Chen$^{2}$, Chen Chen$^{1}$, Ziyan Wu$^{2}$\\[0.2cm]
    $^{1}$University of Central Florida, Orlando, FL, USA\\
    $^{2}$United Imaging Intelligence, Boston, MA, USA\\
    $^{3}$Northwestern University, Chicago, IL, USA\\[0.2cm]
}

\begin{document}
\maketitle
\begin{abstract}

Temporal awareness is essential for video large language models (LLMs) to understand and reason about events within long videos, enabling applications like dense video captioning and temporal video grounding in a unified system. However, the scarcity of long videos with detailed captions and precise temporal annotations limits their temporal awareness. In this paper, we propose \textbf{\method}, a data-oriented training paradigm that leverages sequences of images and short video clips to enhance temporal awareness in long videos. By converting sequence positions into temporal annotations, we transform large-scale image and clip captioning datasets into sequences that mimic the temporal structure of long videos, enabling self-supervised training with abundant time-sensitive data. To enable sequence-to-time knowledge transfer, we introduce a novel time representation that unifies positional information across image sequences, clip sequences, and long videos. Experiments demonstrate the effectiveness of our method, achieving a 27.6\% improvement in F1 score and 44.8\% in CIDEr on the YouCook2 benchmark and a 14.7\% increase in recall on the Charades-STA benchmark compared to the baseline.

% Existing video instruction tuning datasets with temporal annotations lag behind image-based datasets in both quality and quantity, limiting the time-awareness capabilities of video-based LLMs. To address this, we construct a suite of large-scale and well-annotated datasets based on high-quality image captioning and video datasets. Moreover, we updated the existing time-sensitive instruction tuning dataset by introducing the intrinsic hierarchy of videos.  Additionally, we propose a novel time representation, the \textit{relative time token}, to enhance LLMs' temporal understanding abilities. Our contributions lead to improved performance in time-sensitive video tasks, as demonstrated through extensive experiments.

% 1. problem: lack of time-sensitive video data
% 2. we propose a new training paradigm to manually add more time annotation and caption with high quality

\blfootnote{\noindent
\textsuperscript{*} This work was carried out during the internship of Andong Deng at United Imaging Intelligence, Boston MA, USA. \textsuperscript{$\dagger$} Corresponding author.
}

\end{abstract}    
\section{Introduction}

\begin{figure*}[ht]
\centering
% \vspace{-0.2cm}
\scalebox{1}{\includegraphics[width=1 \textwidth]{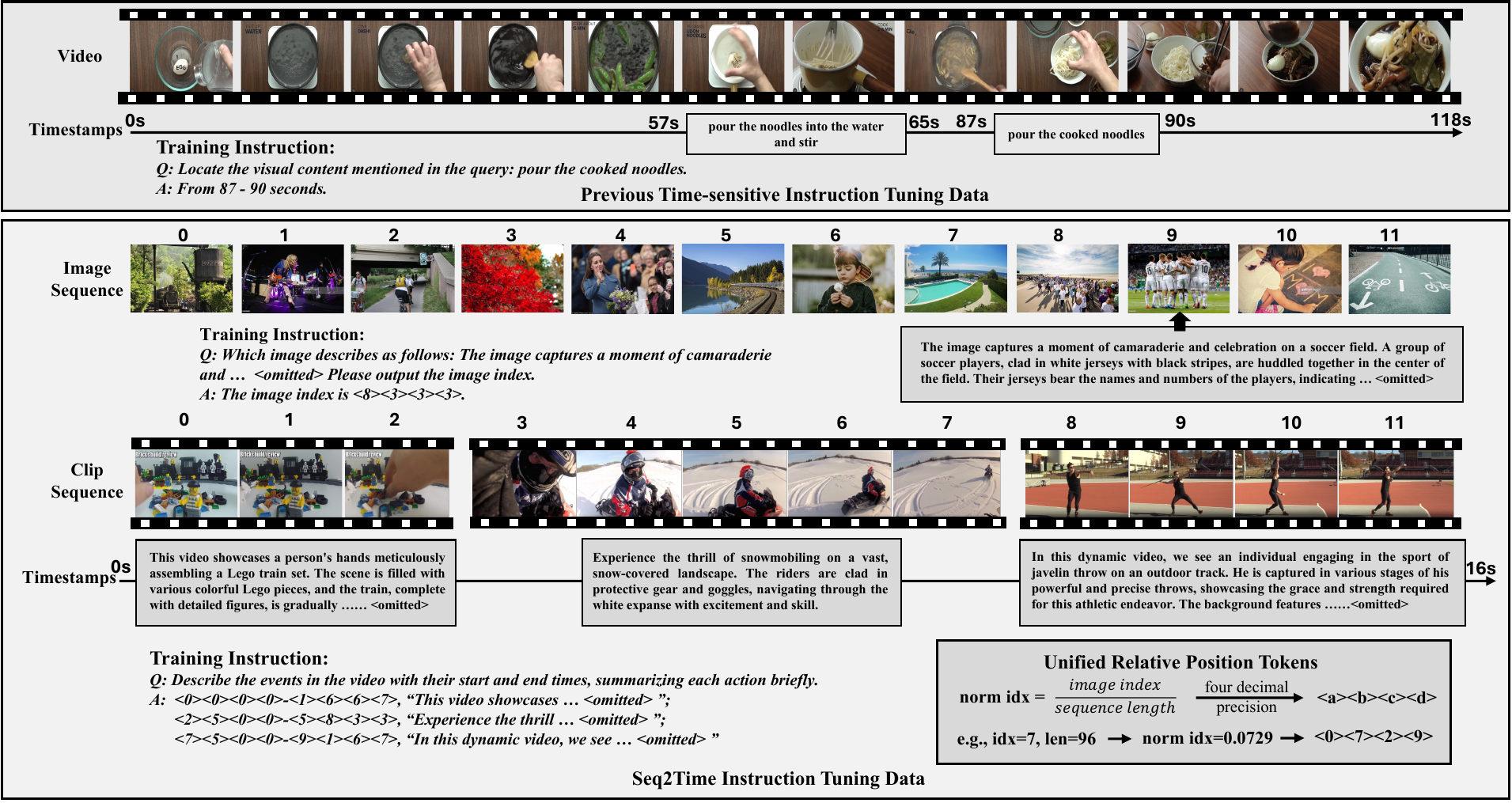}}
\caption{Illustration of our proposed \method. While existing approaches~\cite{ren2024timechat,guo2024vtgllm} rely on timestamp-annotated long videos, Seq2Time leverages image and short video datasets. We introduce a novel time representation of \textbf{unified relative position token} that bridges different sequence types by encoding positions as 4-digit codes---each digit becomes a learnable embedding in the language space, enabling seamless knowledge transfer in the LLM embedding space.}
\label{fig:teaser}
% \vspace{-12pt}
\end{figure*}

% 1. current videoLLM not good.
% scale up
% less existing data (60K timechat)

Developing versatile video large language models (LLMs) capable of perceiving temporal dynamics in long-term videos is crucial for advancing toward artificial general intelligence and has numerous impactful real-world applications, \eg, enabling intelligent surveillance systems to detect anomalous activities, assisting in medical diagnostics through the analysis of patient behavior, and facilitating advanced sports analytics by recognizing patterns and strategies over entire games.

Building upon well-trained LLMs \cite{zhang2022opt,chung2024flant5,dubey2024llama,bai2023qwen,jiang2023mistral}, visual instruction tuning \cite{liu2023llava}—which projects visual representations into the language embedding space—has significantly advanced the ability to perform various multimodal tasks, such as image captioning \cite{zhuminigpt,li2023blip2,liu2023improvedllava}, video question answering \cite{sevila,Maaz2023VideoChatGPT,li2023videochat,damonlpsg2024videollama2,li2024mvbench}, and 3D perception \cite{li2024llava_next_interleave,hong20233dllm,yu2024crema}. The success of multimodal LLMs (MLLMs) in general vision-language tasks has further inspired the development of time-sensitive MLLMs \cite{ren2024timechat,huang2024vtimellm,qian2024momentor,wang2024hawkeye,huang2024lita,guo2024vtgllm} for long-term video understanding, such as dense video captioning \cite{krishna2017dense} and temporal video grounding \cite{gao2017tall,anne2017localizing}.

Creating time-sensitive instruction tuning data from existing video datasets with timestamp annotations \cite{krishna2017dense,gao2017tall,anne2017localizing,zhou2018youcook2,tang2019coin} is essential for temporal grounding. However, current time-related video datasets face significant limitations. First, they are limited in scale due to the labor-intensive process of timestamp annotation; datasets like TimeIT \cite{ren2024timechat} and VTG-IT \cite{guo2024vtgllm} contain only around 125K and 120K videos respectively, which is substantially fewer than general video instruction tuning datasets without detailed time annotation, \eg, VideoChat2 \cite{li2024mvbench} includes over 800K videos. However, the number of training data is crucial in time-sensitive MLLM training. As illustrated in Table~\ref{data_size_dvc}, when randomly removing 12\% task-irrelevant training data for TimeChat~\cite{ren2024timechat}, the average performance of dense video captioning on YouCook2 could drop 13.4\%; while intentionally removing task-relevant training data (\eg, dense video captioning), the performance sharply degrades 65.5\%. Second, the captions in these datasets often lack richness and precision; the visual-language alignment is less accurate compared to high-quality datasets. For example, LLaVA-Instruct-150K \cite{liu2023llava} provides detailed image captions and specifies key object locations, greatly enhancing perception during instruction following. These limitations—the constrained dataset sizes and suboptimal caption quality—not only hinder the training of temporally aware video LLMs but also impede their performance in understanding and reasoning over long-term videos.

To address these challenges, we propose \textbf{\method}, a data-oriented training paradigm that leverages sequences of independent images and short video clips to enhance temporal grounding in long videos. Our intuition is that MLLMs do not inherently ``perceive time" but recognize the correspondence between visual content and its position in a sequence. Therefore, we construct two types of data: i) image sequence data and ii) short video clip sequence data. For the image sequence data (IS), based on correlations between image captions and their indices in the sequence, we design three sophisticated pretext tasks: a) image index grounding, b) indexed image captioning and c) adjacent location reasoning, enforcing the MLLM to link the sequential location of a specific image to its textual content and vice versa. For the clip sequence data (CS), we first generate captions using LongVA \cite{zhang2024longva} given the video and its action labels; then, we combine several short clips along with their sequential positions to form a long video. Similarly, the model is trained to learn the correspondence between the captions of the short clips and their sequential positions. Due to the large scale of image \cite{li2024llavanext-ablations} and short video clip \cite{carreira2019k700} datasets, we can easily sample a vast number of long sequences with high-quality caption annotations, significantly improving the quality and quantity of the instruction tuning training data.

% different advantages for image sequence and clip sequence
% It is worth noting that the image sequence data and the clip sequence data in \method have their own advantages. The three pretext tasks of image sequence are much more challenging than the two downstream tasks, since it is related to locate one or several few images in a long sequence. And the image caption quality is better than the video data. On the other hand, the clip sequence is more aligned with real long videos from both data perspective and training task perspective. In geneal, the two different data is supplementary in our \method.
It is worth noting that the image sequence data and clip sequence data in \method offer distinct advantages. The three pretext tasks associated with image sequences are more challenging than the two downstream tasks, as they require locating one or few specific images within a long sequence. Additionally, the quality of image captions is generally higher than that of video data. On the other hand, the clip sequence data aligns more closely with real-world long video sequences, both in terms of data characteristics and training objectives. Overall, the two types of data are complementary in our \method.

Importantly, to facilitate seamless knowledge transfer from image and clip sequence to temporal understanding in real real-world settings, we introduce a novel time representation called the \textit{unified relative position token}. This representation unifies positional information across image sequences, clip sequences, and long videos within the LLM embedding space. By encoding the relative positions of images or clips as tokens, we enable the model to generalize the concept of position in a sequence to temporal locations in videos. This unification allows the video LLMs to interpret positional cues consistently across different modalities, effectively bridging the gap between artificial sequences and continuous video content. 
% Furthermore, we enhance existing instruction tuning data by introducing the intrinsic hierarchical structure of long videos. Specifically, for instructions with step-wise annotations, we utilize GPT-4 to reorganize the data into step-level, event-level, and video-level descriptions \cite{islam2024videorecap}. Consequently, MLLMs can learn structured knowledge as additional temporal context. 
% During training, we adopt a mixed training and fine-tuning paradigm and utilize the \textit{unified relative position token} instead of free-form timestamp descriptions to unify the representation of image indices and timestamps in the LLM embedding space \zgnote{repetition of previous sentences}.
% Our contributions are summarized as follows: 
% \begin{itemize}[leftmargin=0.15in] \setlength\itemsep{-0.2em} 
% \item We propose \method, a data-oriented training paradigm that leverages sequences of images and short video clips to enhance temporal grounding in long videos. 
% \item We carefully design three pretext tasks for the image sequence data to enable the localization ability of MLLMs.
% \item We introduce a novel time token to unify the representation of image indices in image sequence data and timestamps in video data in the LLM embedding space, facilitating seamless knowledge transfer from sequences to temporal understanding. 
% \item We experimentally demonstrate the effectiveness of our method, improving 44.8\% on CIDEr and 27.6\% on F1 score on YouCook2 and 17.1\% on Recall on Charades-STA for the baseline model. 
% \end{itemize}
Our contributions are summarized as follows: 
\begin{itemize}[leftmargin=0.15in] 
\item We propose \textbf{\method}, a novel data-oriented training paradigm that enhances temporal understanding in video LLMs through image and clip sequences, enabling effective temporal grounding in long videos without requiring extensive timestamp annotations.
\item We introduce three well-designed pretext tasks for image sequence data (image index grounding, indexed image captioning, and adjacent location reasoning) that effectively enhance MLLMs' ability to locate and describe temporal events in sequences.
\item We develop a unified relative position token representation that bridges image indices and video timestamps in the LLM embedding space, facilitating efficient knowledge transfer between different sequence types and temporal understanding tasks.
\item Experiments validate the effectiveness of Seq2Time, achieving significant improvements over the baseline model: 44.8\% gain in CIDEr and 27.6\% in F1 score on YouCook2, and 17.1\% improvement in R@1 (IoU=0.7) on Charades-STA.
\end{itemize}

\begin{table}[t]
    \centering
    \scalebox{1.0}{ % Adjust the scale factor as needed
        \begin{tabular}{lcccc}
            \toprule
            Training Data & \textbf{SODA\_c} & \textbf{CIDEr} & \textbf{Meteor} & \textbf{F1} \\
            \midrule
            Full & 1.0 & 2.9 & 1.1 & 12.7 \\        
            w/o irr & 1.2 & 2.0 & 0.8 & 10.5  \\
            w/o dvc & 0.4 & 0.9 & 0.3 & 4.8 \\
            \bottomrule
        \end{tabular}
    }
    \caption{Impact of data loss on the performance of TimeChat on dense video captioning (YouCook2). We compare the full training data results and 12\% less training data. ``irr" means task-irrelevant data, and ``dvc" means dense video captioning data. The average performance drops 13.4\% and 65.5\%, respectively. }
\label{data_size_dvc}
\vspace{-1em}
\end{table}
\section{Related Work}
\subsection{Visual Instruction Tuning Data}
The emergence of multimodal LLMs \cite{liu2023llava,zhuminigpt,li2023videochat,damonlpsg2024videollama2,Maaz2023VideoChatGPT} has revolutionized vision-language understanding, largely due to advances in visual instruction tuning \cite{liu2023llava} and carefully curated instruction datasets \cite{li2023videochat,liu2023llava,li2024llava_next_interleave,chen2024sharegpt4video}. A significant milestone was established in \cite{liu2023llava}, where they introduced a standardized protocol using GPT-4~\cite{openai2023gpt4} to transform COCO dataset~\cite{lin2014microsoft} annotations into LLM-compatible instruction data. This approach was later extended to video understanding, with \cite{li2023videochat} leveraging WebVid-10M~\cite{Bain21webvid} to create comprehensive video-centric instruction tuning datasets. % LISA~\cite{lai2024lisa} introduced ReasonSeg for reasoning-based segmentation.

Several efforts have been made to create time-sensitive instruction-tuning datasets. LITA~\cite{huang2024lita} developed specialized instruction data for temporal localization. TimeIT~\cite{ren2024timechat} focused on dense video captioning and grounding. However, these time-sensitive datasets, particularly TimeIT, face limitations in both scale and quality due to the resource-intensive nature of timestamp annotation. Our work addresses these challenges by introducing a novel approach that leverages images and short clips to create rich instruction tuning data, as illustrated in Figure~\ref{fig:teaser}, ultimately enhancing video LLMs' temporal understanding capabilities.

\subsection{Time-sensitive MLLMs}
Recent works have made significant strides in developing time-sensitive MLLMs through architectural innovations. TimeChat \cite{ren2024timechat} introduces a dual Q-Former approach: one for timestamp embedding through cross-modal attention and another utilizing sliding windows for long video sequences. VTimeLLM \cite{huang2024vtimellm} employs a three-stage pipeline that progressively learns visual-text alignment, event boundaries, and timestamp perception. Momentor~\cite{qian2024momentor} emphasizes the significance of temporal representation and segment-level modeling, while HawkEye~\cite{wang2024hawkeye} utilizes coarse-grained clip descriptions as intermediate representations with recursive temporal grounding. VTG-LLM~\cite{guo2024vtgllm} introduces absolute-time tokens to minimize quantization errors in time perception, alongside slot-based token compression to address context length limitations. Grounded-VideoLLM~\cite{wang2024grounded} combines relative time tokens with a two-stream visual encoding architecture for improved temporal understanding. While these approaches focus on architectural innovations, our work takes a fundamentally different approach by investigating the potential of \textbf{sequence learning from images and short clips}. We address the critical challenge of insufficient timestamp annotations in long video datasets by exploring, for the first time, how sequential knowledge can be effectively transferred to enhance temporal understanding in long videos.

\subsection{Time-aware Video Understanding}
%TODO
Dense video captioning~\cite{krishna2017dense} and temporal video grounding~\cite{gao2017tall,anne2017localizing} represent tasks in video understanding that require precise temporal comprehension. Recent advances in dense video captioning have explored various innovative approaches: Zhu et al. \cite{zhu2022end} developed an end-to-end framework that integrates ASR signals with simultaneous timestamp and caption prediction, while Yang et al. \cite{yang2023vid2seq} introduced a specialized token-based temporal framework trained on large-scale data. To enhance efficiency and modularity, Zhu et al.~\cite{zhou2024streamdvc} proposed a clustering-based memory bank that can be readily integrated into existing architectures.
While unified frameworks~\cite{yan2023unloc,lin2023univtg} have emerged to handle multiple temporal understanding tasks simultaneously, these approaches are typically designed for specific video understanding tasks with dedicated architectures. In contrast, our work focuses on enhancing Video LLMs' temporal understanding capabilities while preserving their general-purpose reasoning abilities, enabling a more versatile and comprehensive solution for video understanding.

\section{\method}
Traditional approaches to training time-sensitive video LLMs have been constrained by the limited availability of temporal annotations in long video datasets. To address this challenge, we propose \method, a novel paradigm that leverages abundant image and short video data to create rich sequential training data with self-constructed temporal information. Our approach consists of three components: image sequence data (Section~\ref{img_seq}) that exploits index-caption correspondence in long sequences, clip sequence data (Section~\ref{clip_seq}) that utilizes LongVA~\cite{zhang2024longva} to generate high-quality captions for short video clips, and a unified relative position token (Section~\ref{tok}) that bridges different types of sequential data in the LLM embedding space.

\begin{figure*}[ht]
\centering
% \vspace{-0.2cm}
\scalebox{1}{\includegraphics[width=0.96\textwidth]{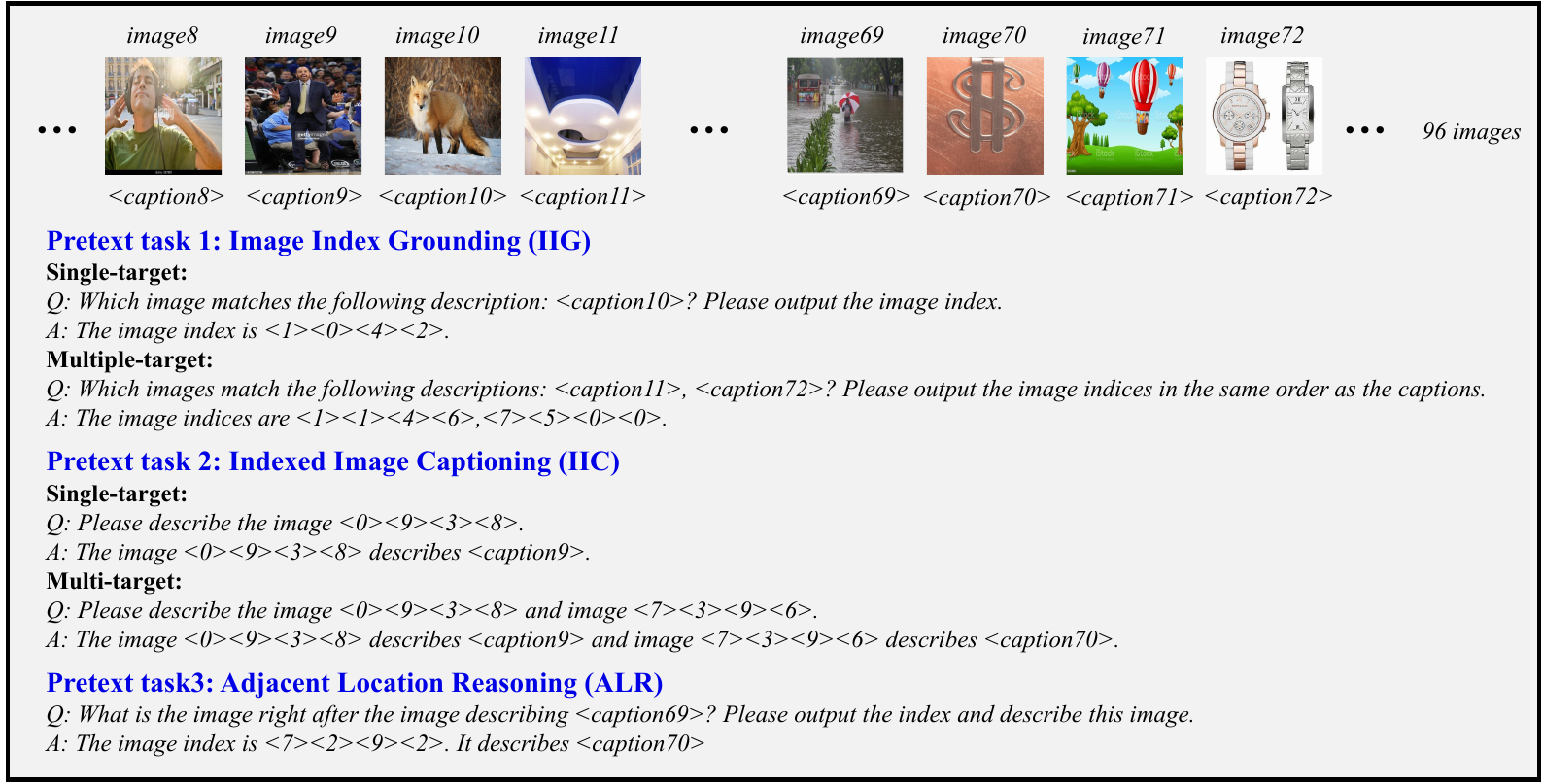}}
\caption{Image sequence data in Seq2Time, featuring three complementary pretext tasks designed to leverage index-caption correspondence. IIG (Image Index Grounding) mimics temporal grounding through position prediction, IIC (Indexed Image Captioning) parallels dense video captioning, and ALR (Adjacent Location Reasoning) enhances sequential understanding through neighbor relationships.}
\label{fig:img_seq}
% \vspace{-12pt}
\end{figure*}

\subsection{Image Sequence Data}
\label{img_seq}
We utilize three LLaVA-ReCap datasets~\cite{li2024llavanext-ablations} (COCO118K, BLIP558K, and CC3M) as our data source due to their rich textual content and consistent image-text alignment. To adapt the image data for downstream time-aware video understanding tasks, we design three pretext tasks: Image Index Grounding, Indexed Image Captioning, and Adjacent Location Reasoning, as shown in Figure~\ref{fig:img_seq}, aimed at enhancing sequence localization abilities in Video LLMs.

\textbf{Image Index Grounding (IIG)} trains the model to locate specific images within a sequence based on their descriptions. Given one or several captions, the model must identify the corresponding image index or indices, mimicking temporal grounding in videos. The input format is: \textit{Which image matches the description: \texttt{<CAPTION>}? Please output the image index}, with output: \textit{The image index is \texttt{<INDEX>}}.

\textbf{Indexed Image Captioning (IIC)} reverses this process by requiring the model to generate descriptions for specific indexed images, analogous to dense video captioning. A typical input is: \textit{Please describe the image with index \texttt{<INDEX>}}, with output: \textit{The image with index \texttt{<INDEX>} describes \texttt{<CAPTION>}}.

\begin{figure*}[ht]
\centering
% \vspace{-0.2cm}
\scalebox{0.95}{\includegraphics[width=1.0\textwidth]{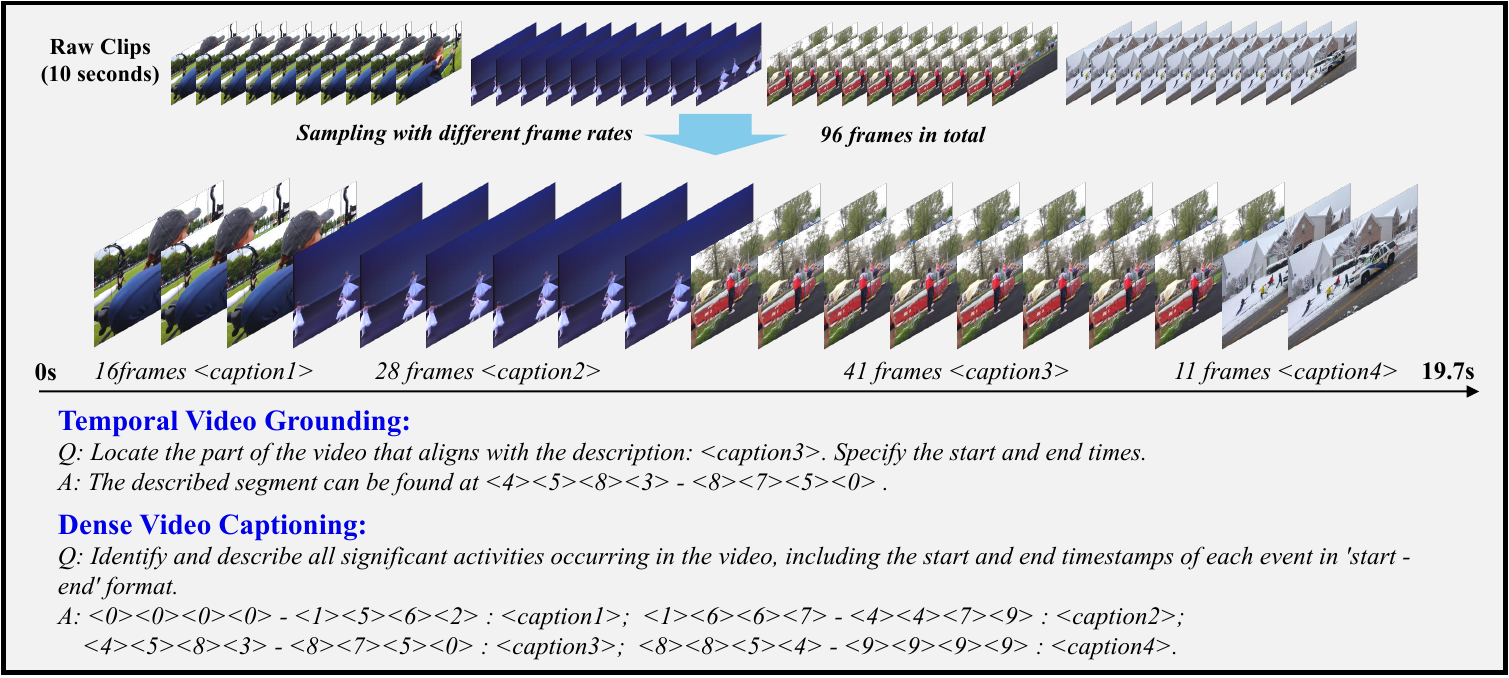}}
\caption{Clip sequence data in Seq2Time. We use LongVA to generate captions for short video clips from Kinetics-700, then combine multiple clips from different action categories to simulate longer videos. Temporal annotations are derived from sequence positions of clips. The resulting sequences serve as training data for both temporal grounding and dense video captioning tasks.}
\label{fig:clipseq}
% \vspace{-12pt}
\end{figure*}

\textbf{Adjacent Location Reasoning (ALR)} challenges the model to understand sequential relationships by identifying and describing neighboring images. The task takes input: \textit{What is the image right before/after the image described as \texttt{<CAPTION1>}? Please provide the index and describe this image}, and expect output: \textit{The image index is \texttt{<INDEX>}. It describes \texttt{<CAPTION2>}}. By combining both index grounding and image description in a sequential context, this task strengthens the model's ability to perform both temporal localization and content understanding, which are essential for video understanding tasks like dense captioning and temporal grounding.

In our implementation, we limit the maximum number of target images to 5 for both IIG and IIC tasks, with 10 different template variations for questions and answers (detailed templates can be found in \textbf{Appendix~\ref{sec:data preparation}}). Each training sequence consists of 96 randomly sampled images, from which up to 5 are selected as targets for caption-based prompts. From our source datasets containing about 3.7M images, we randomly sample 300K training instances, evenly distributed across the three pretext tasks (100K each). This approach not only ensures rich textual descriptions to enhance text generation capabilities but also strengthens sequence localization abilities through our carefully designed pretext tasks.

\subsection{Clip Sequence Data}
\label{clip_seq}
To better simulate real long videos with multiple events, we utilize the Kinetics-700~\cite{carreira2019k700} dataset to construct clip sequence data. We employ LongVA~\cite{zhang2024longva} to generate detailed descriptions for each clip, conditioning on the action labels to ensure high-quality captions. The captioned clips are then combined to create longer sequences, which serve as training data for both dense video captioning and temporal video grounding tasks.
To create diverse temporal structures, we randomly sample between 2 to 10 clips per sequence, mimicking multi-event scenarios in long videos. While most Kinetics-700 clips are 10 seconds long, we intentionally avoid uniform temporal spacing by applying varying frame rates during sampling. This strategy, illustrated in Figure~\ref{fig:clipseq}, ensures that our model learns to handle non-trivial temporal relationships and variable-length segments within sequences.

\subsection{Unified Relative Position Tokens}
\label{tok}
Previous works have represented time in video LLMs using either free-form language (\eg, ``2.05 seconds") \cite{ren2024timechat} or special learnable tokens (\eg, \texttt{<T\_TWO><T\_DOT><T\_ZERO><T\_FIVE>}) \cite{guo2024vtgllm}. However, learning absolute time is often suboptimal for two reasons: 1) videos may have different frame rates, making absolute time less meaningful across datasets, and 2) the primary goal is to capture the correspondence between visual content and its relative sequential position. Furthermore, to enable effective sequence-to-time knowledge transfer, we need to bridge the semantic gap between image indices and video timestamps through a unified representation.

We propose using relative position tokens that work seamlessly across both image sequences and video data (Figures~\ref{fig:img_seq} and {fig:clipseq}). Instead of using direct indices which could lead to complex token sequences (\eg, \texttt{<1><2>}...\texttt{<99>}...), we normalize each position (\ie, image or frame index) to a 4-digit decimal representation: \({I_{\text{norm}}} = \text{round}\left(\frac{i}{L}, 4\right)\), with $i$ the image (or frame) index and $L$ the sequence length. For example, in a 96-image sequence, the 7th image's normalized index would be 0.0729, encoded as \texttt{<0><7><2><9>} (Figure~\ref{fig:teaser}).
The 4-digit precision is carefully chosen to balance accuracy and complexity. For a typical 1-minute video at 30 fps sampled at 96 frames, this representation achieves an average temporal error of just 0.13\%. More specifically, each digit serves a different purpose: the first digit represents the coarse position, while subsequent digits provide increasingly fine-grained localization within that section. This hierarchical structure helps the model learn temporal relationships at different scales.

Our approach introduces only 10 new tokens (\texttt{<0>} through \texttt{<9>}) to the LLM vocabulary, making it highly efficient for training while providing a unified representation that bridges image sequences and video timestamps in the LLM embedding space. During inference, absolute timestamps can be easily reconstructed from these relative positions using the video frame rate, enabling precise temporal localization without the need for learning video-specific timing patterns.

% We propose to use a unified representation for the image index in image sequence data and the timestamps in clip sequence data, which could reduce the learning gap between different sequence formats, thus enhancing the time perception in the downstream tasks.

\section{Experiments}

\begin{table*}[h]
    \centering
    \resizebox{\textwidth}{!}{
    \begin{tabular}{lcccccc}
        \toprule
        \multirow{2}{*}{\textbf{Models}} & \multicolumn{4}{c}{\textbf{YouCook2 (DVC)}} & \multicolumn{2}{c}{\textbf{Charades-STA (TVG)}} \\
        \cmidrule(lr){2-5} \cmidrule(lr){6-7}
        & \textbf{SODA\_c $\uparrow$} & \textbf{CIDEr $\uparrow$} & \textbf{Meteor $\uparrow$} & \textbf{F1 $\uparrow$} & \textbf{R@1 (IoU=0.5) $\uparrow$} & \textbf{R@1 (IoU=0.7) $\uparrow$} \\
        \midrule
        % \textit{time instruction tuning model} \\
        VTimeLLM\cite{huang2024vtimellm} (CVPR 2024) & \textbackslash{} & \textbackslash{} & \textbackslash{} & \textbackslash{} & 27.5 & 11.4 \\
        % HawkEye~\cite{wang2024hawkeye} (arXiv 2024) & \textbackslash{} & \textbackslash{} & \textbackslash{} & \textbackslash{} & 31.4 & 14.5 \\
        Monmentor~\cite{qian2024momentor} (arXiv 2024) & \textbackslash{} & \textbackslash{} & \textbackslash{} & \textbackslash{} & 26.6 & 11.6 \\
        % \textit{Temporal Grounding Video LLMs} \\
        TimeChat~\cite{ren2024timechat}* (CVPR 2024) & 1.0 & 2.9 & 1.1 & 12.7 & 27.2 & 11.7 \\
        % VTG-LLM~\cite{guo2024vtgllm} (arXiv 2024) & 1.5 & 5.1 & 1.8 & 17.2 & 33.8 & 15.7 \\
         % TRACE~\cite{guo2024trace} (arXiv 2024) & 2.2 & 8.1 & 2.8 & 22.4 & 40.3 & 19.4  \\
        \midrule
        % \textit{Seq2Time} \\
        % TimeChat + \method (I) & 1.2 & 3.4 & 1.2 & 14.3 & 24.0 & 10.4 \\
        TimeChat + \method (\textbf{Ours}) w/o RPT & 1.2 & 3.7 & 1.4 & 15.7 & 29.3 & 12.8 \\ % LLM
        TimeChat + \method (\textbf{Ours}) & 1.3\textsubscript{\textcolor{red}{\textbf{+30.0\%}}} & 4.2\textsubscript{\textcolor{red}{\textbf{+44.8\%}}} & 1.3\textsubscript{\textcolor{red}{\textbf{+18.2\%}}} & 16.2\textsubscript{\textcolor{red}{\textbf{+27.6\%}}} & 31.2\textsubscript{\textcolor{red}{\textbf{+14.7\%}}} & 13.7\textsubscript{\textcolor{red}{\textbf{+17.1\%}}} \\
        % TRACE + \method (\textbf{Ours}) \\ % LLM
        % VTG-LLM + MT + FT & 1.4 & 4.4 & 1.6 & 18.0 & \textbackslash{} & \textbackslash{} \\
        % VTG-LLM + MT + FT* & 1.6 & 4.9 & 2.0 & 16.8 & \textbackslash{} & \textbackslash{} \\
        \bottomrule
    \end{tabular}
    }
    \caption{Comparison on YouCook2 and Charades-STA datasets. * denotes our reproduction of TimeChat with partial training data availability, resulting in lower scores than reported~\cite{ren2024timechat}. RPT denotes our proposed unified relative position token for time representation.} % Seq2Time demonstrates consistent improvements across both benchmarks, with particularly notable gains in YouCook2's CIDEr (+44.8\%) and F1 scores (+27.6\%).
\label{tab:main}
\vspace{-5pt}
\end{table*}

\subsection{Experiment Setup}
\noindent\textbf{Training Data}\quad Our training dataset comprises three main components: 1) Seq2Time instruction data, consisting of 300K image sequence instances and 100K clip sequence instances; 2) 110K temporal understanding samples from TimeIT~\cite{ren2024timechat}; and 3) general video caption data, including 40K instances from Valley~\cite{luo2023valley} and 93K from ShareGPT4Video~\cite{chen2024sharegpt4video}. For Seq2Time data, we employ balanced sampling across tasks: the image sequence tasks (IIG, IIC, and ALR) are randomly selected with equal probability; and similarly, the clip sequence tasks (dense video captioning and temporal video grounding) are sampled uniformly.\\

\noindent\textbf{Implementation Details}\quad We evaluate \method using TimeChat~\cite{ren2024timechat} as our baseline model. The training process consists of two phases: initial training for 1 epoch with the complete dataset, followed by 3 epochs of fine-tuning using only TimeIT and Valley (40K) data. Training parameters include a LoRA rank of 32, batch size of 8, and 96 sampled frames per video, following TimeChat's configuration. All experiments are conducted on 8 A100 GPUs (80GB each).
For evaluation, we follow TimeChat's metrics: dense video captioning is assessed using \textit{SODA\_c}, \textit{CIDEr}, and \textit{Meteor} for text quality, and \textit{F1 score} for temporal accuracy. Temporal video grounding is evaluated using \textit{R@1} with IoU thresholds of 0.5 and 0.7. Note that our reproduced baseline results are slightly lower than those reported in~\cite{ren2024timechat} due to partial Valley~\cite{luo2023valley} data availability.

\subsection{Main Results}
We evaluate \method on two standard benchmarks for temporal understanding: YouCook2 for dense video captioning and Charades-STA for temporal video grounding (Table~\ref{tab:main}). Our approach demonstrates consistent improvements across all metrics compared to the TimeChat baseline, even without the unified relative position token (RPT). The improvements are particularly notable in YouCook2's CIDEr and F1 scores, and Charades-STA's R@1(IoU=0.5), validating that our sequence-based training effectively enhances temporal understanding capabilities.

The addition of RPT further boosts performance, yielding significant improvements: +30\% in SODA\_c and +27.6\% in F1 score on YouCook2, and +17.1\% in R@1(IoU=0.7) on Charades-STA. Notably, while TimeChat relies on approximately 120K manually annotated temporal samples, \method achieves superior performance using largely self-supervised data, leveraging the natural sequential correspondence between images/clips and their captions. We provide detailed ablation studies of individual components in the following section.

% Additionally, we also apply \method on TRACE~\cite{guo2024trace} to investigate the versatility of \method. 
% We also evaluate the viability of \method on TRACE. Since TRACE is trained on a large-scale dataset, due to the computation resource, we did not train from scratch with \method but directly on top of pretrained TRACE. TODO

% \begin{table*}[h!]
%     \centering
%     \begin{tabular}{lcccccc}
%         \toprule
%         \multirow{2}{*}{\textbf{IS/CS}} & \multicolumn{4}{c}{\textbf{YouCook2 (DVC)}} & \multicolumn{2}{c}{\textbf{Charades-STA (TVG)}} \\
%         \cmidrule(lr){2-5} \cmidrule(lr){6-7}
%         & \textbf{SODA\_c} & \textbf{CIDEr} & \textbf{Meteor} & \textbf{F1} & \textbf{R@1(IoU=0.5)} & \textbf{R@1(IoU=0.7)} \\
%         \midrule
%         300K/100K & \cellcolor{darkgreen}1.3 & \cellcolor{darkgreen}4.2 & \cellcolor{darkgreen}1.3 & \cellcolor{darkgreen}16.2 & \cellcolor{darkgreen}31.2 & \cellcolor{darkgreen}13.7  \\
%          60K/20K  & 1.1 & \cellcolor{lightgreen}3.8 & \cellcolor{darkgreen}1.3 & 14.8 & \cellcolor{lightgreen}28.3 & \cellcolor{lightgreen}11.9   \\
%          15K/5K  & \cellcolor{lightgreen}1.2 & 3.6 & \cellcolor{lightgreen}1.2 & \cellcolor{lightgreen}15.0 & 26.8 & 11.2   \\
%          0K/0K  & 1.0 & 2.9 & 1.1 & 12.7 & 27.2 & 11.7   \\
%         \bottomrule
%     \end{tabular}
%     \caption{Seq2Time data scaling results on TimeChat. We can observe a performance jump with only 5\% of \method data, and in general, more data leads to better overall performance.}
%     \label{tab:dataset_scaling}
% \end{table*}

\subsection{Dataset Effect}
To analyze how different data components influence temporal understanding in video LLMs, we conduct comprehensive ablation studies on TimeChat, as shown in Table~\ref{tab:dataset_ablation}. Removing clip sequence (CS) data leads to performance drops in both dense video captioning and temporal grounding tasks, demonstrating its importance in the training pipeline.

\begin{table}[h!]
    \centering
    \setlength{\tabcolsep}{4pt} % Reduces column padding
    \small % Optional: reduces font size
    \begin{tabular}{lcccccccc}
        \toprule
        \multirow{2}{*}{\textbf{IS}} & \multirow{2}{*}{\textbf{MC}} & \multirow{2}{*}{\textbf{CS}} & \multicolumn{4}{c}{\textbf{YouCook2}} & \multicolumn{2}{c}{\textbf{Charades-STA}} \\
        \cmidrule(lr){4-7} \cmidrule(lr){8-9}
        & & & \textbf{S $\uparrow$} & \textbf{C $\uparrow$} & \textbf{M $\uparrow$} & \textbf{F1 $\uparrow$} & \textbf{R@0.5 $\uparrow$} & \textbf{R@0.7 $\uparrow$} \\
        \midrule
        \lgcmark & \lgcmark & \lgcmark & \cellcolor{lightgreen}1.3 & \cellcolor{lightgreen}4.2 & \cellcolor{lightgreen}1.3 & \cellcolor{darkgreen}16.2 & \cellcolor{darkgreen}31.2 & \cellcolor{darkgreen}13.7 \\
        \midrule
        \lgcmark & \lgcmark & \lgxmark & 1.2 & 4.0 & 1.2 & \cellcolor{lightgreen}15.9 & \cellcolor{lightgreen}30.9 & \cellcolor{lightgreen}13.1 \\
        \lgcmark & \lgxmark & \lgxmark & \cellcolor{darkgreen}1.4 & \cellcolor{darkgreen}4.3 & \cellcolor{darkgreen}1.4 & 13.3 & 28.8 & 12.5 \\
        \lgxmark & \lgxmark & \lgxmark & 1.0 & 2.9 & 1.1 & 12.7 & 27.2 & 11.7  \\
        \bottomrule
    \end{tabular}
    \caption{Dataset ablation of Seq2Time on TimeChat.  IS: image sequence data, MC: more video captions, CS: clip sequence data. Dark green and light green indicate the best and second-best results, respectively, and this color scheme applies to all tables unless stated otherwise.}
    \label{tab:dataset_ablation}
\end{table}

\begin{table}[h!]
    \centering
    \setlength{\tabcolsep}{4pt} % Reduces column padding
    \renewcommand{\arraystretch}{0.8} % Reduces row height
    \small
    \begin{tabular}{lcccc}
        \toprule
        % \multirow{2}{*}{\textbf{Video LLaMA}} & \multicolumn{4}{c}{\textbf{YouCook2}} \\
        % \cmidrule(lr){2-5} 
        \textbf{Video LLaMA} & \textbf{S $\uparrow$} & \textbf{C $\uparrow$} & \textbf{M $\uparrow$} & \textbf{F1 $\uparrow$} \\
        \midrule
        w/o IS & 0.1 & 0.0 & 0.0 & 0.2  \\        
        w/ IS & 0.1 & 0.1 & 0.2 & \textbf{3.3}  \\
        \bottomrule
    \end{tabular}
    \caption{Impact of image sequence data on Video-LLaMA. Training with only image sequence data enables temporal awareness as demonstrated by improved F1 scores (0.2 → 3.3).}
    \label{tab:is_on_dvc}
\vspace{-5pt}
\end{table}

Interestingly, incorporating additional video captions (MC) from ShareGPT4video~\cite{chen2024sharegpt4video} shows unexpected effects: while it doesn't significantly improve captioning quality, it substantially enhances temporal perception, yielding a 19.5\% improvement in YouCook2's F1 score and 7.3\% in Charades-STA's R@1(IoU=0.5). We hypothesize that exposure to more captions improves the model's ability to understand long-form video content, leading to better event boundary detection. The impact of image sequence data (IS) is particularly striking, especially in text generation quality. The CIDEr score shows a remarkable 48.3\% improvement over the baseline, providing strong evidence that temporal understanding can be effectively learned from static image sequences. Notably, while the addition of MC and CS improves temporal localization, it results in a slight degradation in text generation quality. This may be because image sequence data inherently provides stronger sequential visual-language alignment (\ie per-image captions), whereas video caption data represents visual-language correspondence at the video level. Additionally, the limitations of LongVA~\cite{zhang2024longva} may also impact the quality of clip-level captions, further contributing to this effect.

To further validate the effectiveness of the image sequence data, we tested our image sequence data on Video-LLaMA~\cite{damonlpsg2024videollama2}, a general video LLM without prior temporal training. As shown in Table~\ref{tab:is_on_dvc}, while the base model struggles with temporal events (F1 score of 0.2), adding image sequence (w/ IS) improves this to 3.3, showcasing initial temporal awareness. Notably, this improvement occurs even without our unified relative position tokens, using only free-form language for index representation. This confirms that sequential knowledge can be effectively transferred across different data modalities.

In addition to dense video captioning performance with common metrics, we also evaluate the impact of different training datasets on the text richness of the generated captions in two aspects: generated caption length and lexical diversity. As shown in Table~\ref{tab:ttr}, adding new training data has a noticeable effect on the text richness of the captions. To assess text richness, we measure the average caption length ($L_{\text{avg}}$) and the Type-Token Ratio (TTR)~\cite{malvern2004lexical}, which quantifies lexical diversity. As indicated in the results, incorporating additional training data consistently enhances the richness and diversity of the generated captions, resulting in more descriptive and meaningful outputs.

\begin{table}[t]
\centering
\small
\begin{tabular}{lccccc}
\toprule
\textbf{Method}  & \textbf{TimeChat} & \textbf{+IS} & \textbf{+IS+MC} & \textbf{+\method} \\ 
\midrule
$L_{\text{avg}}$  & 6.81 & 8.19 &   \cellcolor{lightgreen}8.74   &  \cellcolor{darkgreen}11.53 \\ 
TTR      & 0.268   & \cellcolor{darkgreen}0.411     &  0.384       &  \cellcolor{lightgreen}0.396  \\ 
\bottomrule
\end{tabular}
\caption{Impact of caption quality metrics across different training data configurations. $L_{\text{avg}}$: average caption length, TTR: Type-Token Ratio (lexical diversity). Higher values indicate more detailed and diverse descriptions.}
\label{tab:ttr}
\vspace{-5pt}
\end{table}

% This indicates learning the image-captioning correspondence from a long image sequence could indeed enhance the time-sensitivity of video LLMs.

\subsection{Image-based Pretext Tasks Ablation}
% In this section, we experimentally evaluate the individual effect of each pretext task in our image sequence data. In the experiments, to remove the effect of the number of training samples, we fix the total number of image sequence data but only change the existence of a specific pretext task. For instance, in the second row of Table~\ref{tab:pretext_task_ablation}, the adjacent location reasoning is removed during training, and the sampling number of the other two pretext tasks increases to 150K each. Comparing the first two rows, we can observe a clear drop in CIDEr, which indicates the effectiveness of indexed image captioning (IIC) on text generation. After removing the image index grounding (IIG), we can observe a general performance degradation on both benchmarks. This demonstrates that IIG plays a more important role compared with IIC. Additionally, comparing the third row with the last row, which is the TimeChat baseline results, we can also obtain another interesting result that only increasing the number of training data cannot guarantee a better performance, the data composition is a more critical factor. For the adjacent location reasoning (ALR),  comparing the 4-th row with the 1-st row, we can observe that ALR has a significant impact on the text generation performance, after removing this task, the CIDEr drop 37.2\% while SODA\_c drops 40\%. Moreover, comparing the last two rows, we can also see the the sequence localization improvement brought by IIC and IIG, \ie, a 13.4\% improvement after adding 300K IIC and IIG data.

We analyze the contribution of each pretext task while controlling for training data volume by maintaining a fixed total amount of image sequence data. Table~\ref{tab:pretext_task_ablation} shows our systematic evaluation where omitted tasks are compensated by increasing the sampling of remaining tasks (\eg when ALR is removed, IIC and IIG samples increase to 150K each).

\begin{table}[t]
    \centering
    \setlength{\tabcolsep}{3pt} % Adjusts the column padding, you can decrease or increase this value as needed
    \small % Optional: reduces font size
    \begin{tabular}{lcccccccc}
        \toprule
        \multirow{2}{*}{\textbf{IIC}} & \multirow{2}{*}{\textbf{IIG}} & \multirow{2}{*}{\textbf{ALR}} & \multicolumn{4}{c}{\textbf{YouCook2}} & \multicolumn{2}{c}{\textbf{Charades-STA}} \\
        \cmidrule(lr){4-7} \cmidrule(lr){8-9}
        & & & \textbf{S $\uparrow$} & \textbf{C $\uparrow$} & \textbf{M$\uparrow$} & \textbf{F1 $\uparrow$} & \textbf{R@0.5 $\uparrow$} & \textbf{R@0.7) $\uparrow$} \\
        \midrule
        \lgcmark & \lgcmark & \lgcmark & \cellcolor{darkgreen}1.4 & \cellcolor{darkgreen}4.3 & \cellcolor{darkgreen}1.4 & \cellcolor{lightgreen}13.3 & \cellcolor{darkgreen}28.8 & \cellcolor{darkgreen}12.5 \\
        \lgxmark & \lgcmark & \lgcmark & \cellcolor{lightgreen}1.2 & \cellcolor{lightgreen}3.4 & \cellcolor{lightgreen}1.2 & 12.9 & \cellcolor{lightgreen}27.9 & \cellcolor{lightgreen}11.8 \\
        \lgcmark & \lgxmark & \lgcmark & 1.1 & 2.6 & 0.9 & 11.9 & 26.2 & 11.3 \\
        \lgcmark & \lgcmark & \lgxmark & 1.0 & 2.7 & 1.0 & \cellcolor{darkgreen}14.4 & 27.2 & 11.4 \\
        \midrule
        \lgxmark & \lgxmark & \lgxmark & 1.0 & 2.9 & 1.1 & 12.7 & 27.2 & 11.7 \\
        \bottomrule
    \end{tabular}
    \caption{Effect of individual pretext tasks in image sequence training. The first row shows results with only image sequence data. IIC: Indexed Image Captioning, IIG: Image Index Grounding, ALR: Adjacent Location Reasoning.} % \textcolor{darkgreen}{Dark green} and \textcolor{lightgreen}{light green} shading indicates the best and second-best results, respectively (the same in the following tables unless stated otherwise).
    \label{tab:pretext_task_ablation}
    \vspace{-12pt}
\end{table}

Our analysis reveals distinct roles for each task. Indexed Image Captioning (IIC) primarily \textbf{enhances text generation quality}, as evidenced by the significant CIDEr score drop when IIC is absent. Image Index Grounding (IIG) \textbf{proves crucial for overall temporal understanding}, with its removal causing performance degradation across both benchmarks, suggesting a stronger impact than IIC. Adjacent Location Reasoning (ALR) \textbf{substantially influences text generation capabilities}—its removal leads to a 37.2\% decrease in CIDEr and 40\% in SODA\_c scores.

Importantly, comparing the baseline TimeChat with variants using only IIC and IIG shows that performance improvements depend more on task diversity than data volume—adding 300K samples of just IIC and IIG data yields only a 13.4\% improvement. This underscores the importance of our complementary task design in developing robust temporal understanding capabilities.

\begin{figure*}[ht]
\centering
% \vspace{-0.2cm}
\scalebox{1}{\includegraphics[width=1\textwidth]{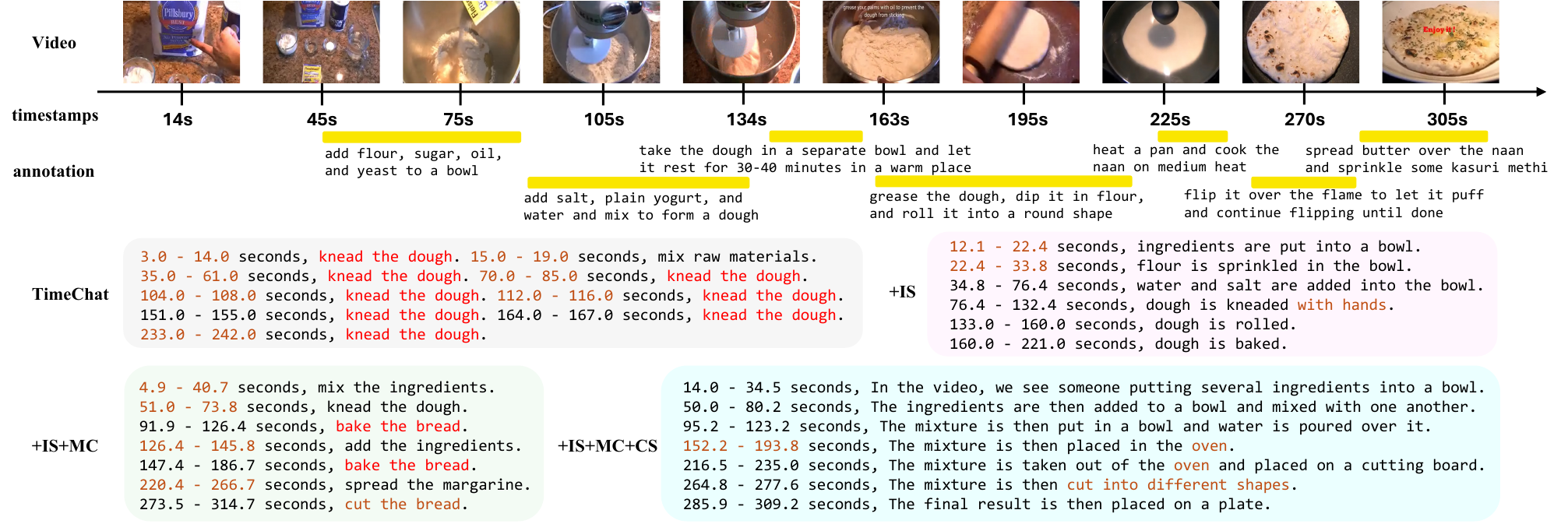}}
\caption{Qualitative examples of our \method on TimeChat. IS: image sequence data, CS: clip sequence data, MC: more video captions. Text highlighted in \textcolor{red}{\textbf{red}} shows repetitive patterns in outputs, while \textcolor{brown}{\textbf{brown}} indicates incorrect predictions in timestamps or event descriptions.}  %  Adding image sequence data significantly reduces TimeChat's tendency toward repetition. While some object recognition errors persist (\eg, confusing pans with ovens), Seq2Time training yields both more accurate event descriptions and more precise temporal predictions.
\label{fig:example}
\vspace{-6pt}
\end{figure*}

\subsection{Dataset Scaling}
To investigate the scaling properties of Seq2Time, we conduct experiments with varying proportions (5\% and 20\%) of our image sequence and clip sequence data using TimeChat as the baseline model. As shown in Table~\ref{tab:dataset_scaling}, even with just 5\% of our data, we observe notable improvements in dense video captioning performance, particularly in the CIDEr metric which measures text generation quality. The performance scales positively with data volume, showing consistent improvements across both dense video captioning and temporal video grounding tasks as we increase the proportion of \method data. 

\begin{table}[h!]
    \centering
    \setlength{\tabcolsep}{3pt} % Adjusts the column padding
    \renewcommand{\arraystretch}{0.8} % Adjusts the row height (default is 1.0)
    \small % Optional: reduces font size
    \begin{tabularx}{0.45\textwidth}{rXXXXXX}
        \toprule
        \multirow{2}{*}{\textbf{IS/CS}} & \multicolumn{4}{c}{\textbf{YouCook2}} & \multicolumn{2}{c}{\textbf{Charades-STA}} \\
        \cmidrule(lr){2-5} \cmidrule(lr){6-7}
        & \textbf{S} & \textbf{C} & \textbf{M} & \textbf{F1} & \textbf{R@0.5} & \textbf{R@0.7} \\
        \midrule
        100\% & \cellcolor{darkgreen}1.3 & \cellcolor{darkgreen}4.2 & \cellcolor{darkgreen}1.3 & \cellcolor{darkgreen}16.2 & \cellcolor{darkgreen}31.2 & \cellcolor{darkgreen}13.7  \\
         20\%  & 1.1 & \cellcolor{lightgreen}3.8 & \cellcolor{darkgreen}1.3 & 14.8 & \cellcolor{lightgreen}28.3 & \cellcolor{lightgreen}11.9   \\
         5\%  & \cellcolor{lightgreen}1.2 & 3.6 & \cellcolor{lightgreen}1.2 & \cellcolor{lightgreen}15.0 & 26.8 & 11.2   \\
         0\%  & 1.0 & 2.9 & 1.1 & 12.7 & 27.2 & 11.7   \\
        \bottomrule
    \end{tabularx}
    \caption{Impact of Seq2Time data scaling on TimeChat. Even with 5\% data, significant improvements are observed. Performance continues to scale positively with increased data volume.}
    \label{tab:dataset_scaling}
\vspace{-8pt}
\end{table}

\subsection{Qualitative Study}

% To qualitatively demonstrate the effectiveness of \method, we present a comparison based on TimeChat in Figure~\ref{fig:example}. The original TimeChat often generates redundant text, highlighted by the red text in the figure, a problem that is significantly mitigated after training with \method. This phenomenon may result from the limited captioning data in TimeIT; however, \method benefits from high-quality captioning data available in both image and clip sequence datasets. Besides, \method could also address the hullucination problem in dense video captioning task. In the second example in Figure~\ref{fig:example}, the original TimeChat wrongly response with "takes a spoon of water from the sink" in the beginning segments, in comparison, after training with \method, the model tends to output correct description that "put a pot of water on the stove".
% Additionally, the output from \method provides more detailed object descriptions as illustrated in the first example. Nevertheless, there are still some failure response, for instance, in the third example, the "tofu" is mistakenly recognized as "tomato". This could be potentially solved with more advanced visual encoder in future study.

To qualitatively demonstrate the effectiveness of Seq2Time, we analyze different model variants in Figure~\ref{fig:example}. The baseline \textbf{TimeChat tends to generate repetitive patterns}, \eg, repeatedly mentioning \textit{knead the dough} throughout the video when this action only occurs between 120s and 200s. This phenomenon is consistent with TimeChat's low TTR in Table~\ref{tab:ttr}. This limitation likely stems from insufficient captioning data in TimeIT. However, our \method, leveraging rich captions from both image and clip sequence datasets, significantly improves temporal and descriptive accuracy.

Specifically, the addition of image sequence data (IS) yields two key improvements. \textbf{First, it eliminates repetitive captioning patterns,} as indicated by the absence of repetitive patterns in the green patch. \textbf{Second, it enhances event recognition}—for example, the model accurately localizes the event \textit{water and salt are added into the bowl} to 34.8-76.4 seconds, closely matching the ground truth of 48-79 seconds. This improvement aligns with our quantitative findings in Table~\ref{tab:dataset_ablation}, where image sequence data substantially improves the CIDEr score.
In the full version (blue patch), the model demonstrates even \textbf{more precise temporal and descriptive capabilities}. For instance, it correctly identifies and localizes the event \textit{the mixture is then put in a bowl and water is poured over it} to 95.2-123.2 seconds, matching the ground truth event \textit{add salt, plain yogurt, and water and mix to form a dough} (80-133 seconds). While some fine-grained object recognition challenges still remain, temporal localization is accurate. More visualizations can be found in \textbf{Appendix~\ref{supp: vis}}.

\noindent \textbf{Failure Case Analysis}\quad Despite overall improvements, we still observe several types of prediction errors. As illustrated in Figure~\ref{fig:example}, with only image sequence data, object recognition errors occur—for example, misidentifying a blender as hands during dough mixing. Adding more video caption data (+IS+MC) reduces but doesn't eliminate all repetitive patterns, and sometimes confuses similar objects (\eg, identifying dough as bread). Even in the full version, we notice object recognition errors (pan mistaken for oven) and hallucinated events (\eg, \textit{cut into different shapes}). These issues primarily stem from visual recognition limitations and suggest that incorporating more advanced visual encoders could further improve model performance.

% The original TimeChat often generates redundant text, highlighted by the red text in the figure, a problem that is significantly mitigated after training with \method. This phenomenon may result from the limited captioning data in TimeIT; however, \method benefits from high-quality captioning data available in both image and clip sequence datasets. Moreover, \method also addresses the hallucination problem in dense video captioning tasks. In the second example in Figure~\ref{fig:example}, the original TimeChat incorrectly responds with "takes a spoon of water from the sink" in the beginning segments. In comparison, after training with \method, the model outputs the correct description, "puts a pot of water on the stove." 
% Additionally, the output from \method provides more detailed object descriptions, as illustrated in the first example. Nevertheless, some errors persist; for instance, in the third example, "tofu" is mistakenly recognized as "tomato." 
\section{Conclusion}
% We propose \textbf{\method} to address the data issue in time-sensitive video LLM training. Specifically, we construct image sequence data and clip sequence data and leverage the visual-caption correspondence in these sequence in a self-supervise manner to construct the training pipeline. To reduce the gap between difference sequence, we further introduce unified relative position token, which unify the position representation in the LLM space. The experiment results suggest the effectiveness of \method, and its scalability also shows the potential. In general, \method could well transfer sequential knowledge from short sequence to real long video, which could enhancing the time-perception of video LLMs. In future study, different sequence could be constructed for similar purpose, \eg, image patch sequence could also be utilized, and the correlation between patches could be potentially utilized as supervision.

We propose \method to address the data scarcity challenge in training time-sensitive video LLMs. Our approach constructs training data from both image sequences and video clips in a self-supervised manner. Furthermore, we propose a unified relative position token, which bridges different sequence types by standardizing position representation in the LLM embedding space. Extensive experiments demonstrate the effectiveness and promising scalability of our approach, showing that sequential knowledge can be successfully transferred from sequence data to long video understanding. We believe our work opens several directions for future research. The framework could be extended to other sequence types, such as image patches, where patch-level correlations could provide additional supervisory signals for temporal understanding. This suggests a broader potential for leveraging various forms of sequential data to enhance the temporal capabilities of video LLMs.
{
    \small \bibliographystyle{ieeenat_fullname}
    \bibliography{main}
}

% WARNING: do not forget to delete the supplementary pages from your submission 
\clearpage
% \setcounter{page}{1}
% \maketitlesupplementary
\section*{Appendix}
\appendix
\section{Data Preparation}
\label{sec:data preparation}

\subsection{Image Sequence}
\begin{figure*}[h]
\centering
% \vspace{-0.2cm}
\scalebox{1}{\includegraphics[width=1 \textwidth]{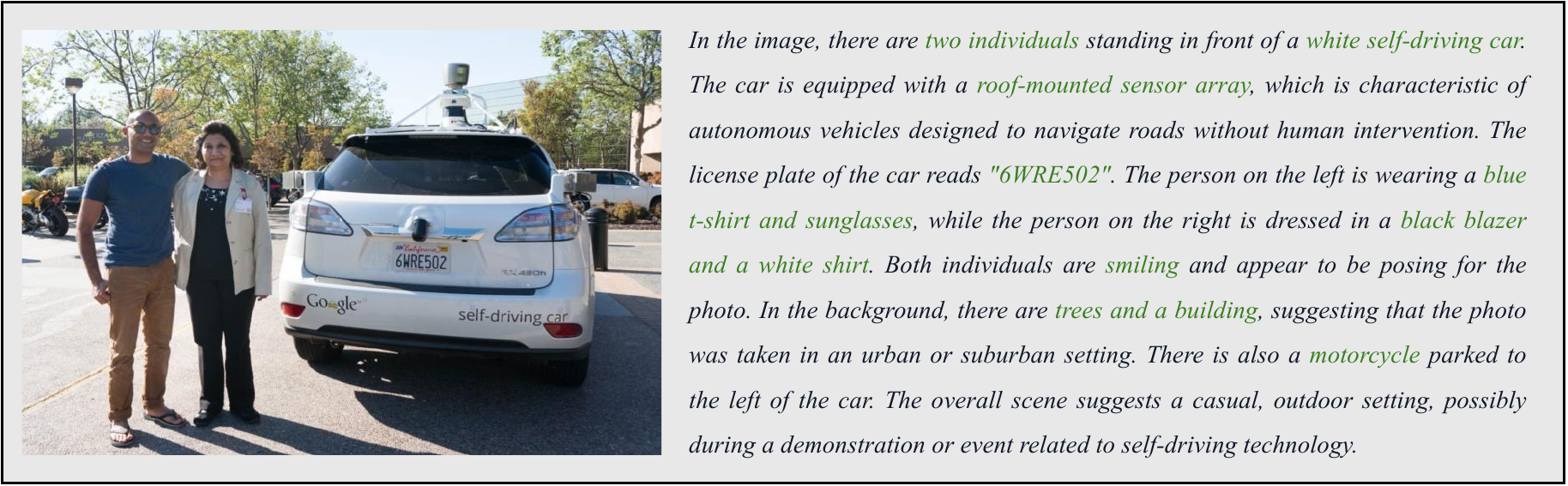}}
\caption{Illustration of an example from LLaVA-ReCap-CC3M. The caption provide details in every aspects of the image from foreground to background. For instance, even the motorcycle in the background is captured. The important descriptions are indicated in \textcolor{darkgreen}{green} texts.}
\label{fig:supp_llava_recap}
% \vspace{-12pt}
\end{figure*}

\noindent\textbf{Data Source.} To construct an image sequence dataset that provides high-quality visual-text alignment, we select LLaVA-ReCap~\cite{li2024llavanext-ablations} as the primary source. Built on the robust vision-language model LLaVA-NeXT-34B~\cite{li2024llava_next_interleave}, LLaVA-ReCap contains over 3.7 million image-text pairs. Its captions are detailed, providing a rich description of the visual content, as exemplified in Figure~\ref{fig:supp_llava_recap}. These captions enable strong correspondence between visual elements and language, serving as a reliable basis for generating high-quality sequential supervision in our training. This alignment is critical for training models to understand and reason over temporally structured visual data.

\begin{figure*}[p]
\centering
% \vspace{-0.2cm}
\scalebox{1}{\includegraphics[width=1 \textwidth]{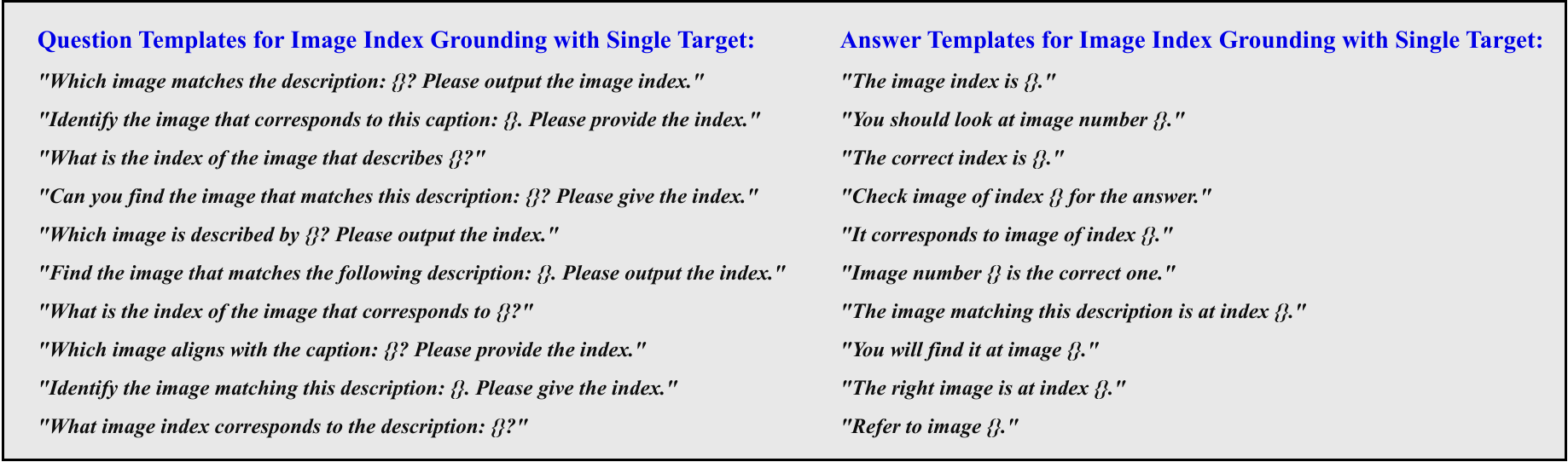}}
\caption{Instruction templates of image index grounding with single target image.}
\label{fig:supp_iig_single}
% \vspace{-12pt}
\end{figure*}

\begin{figure*}[ht]
\centering
% \vspace{-0.2cm}
\scalebox{1}{\includegraphics[width=1 \textwidth]{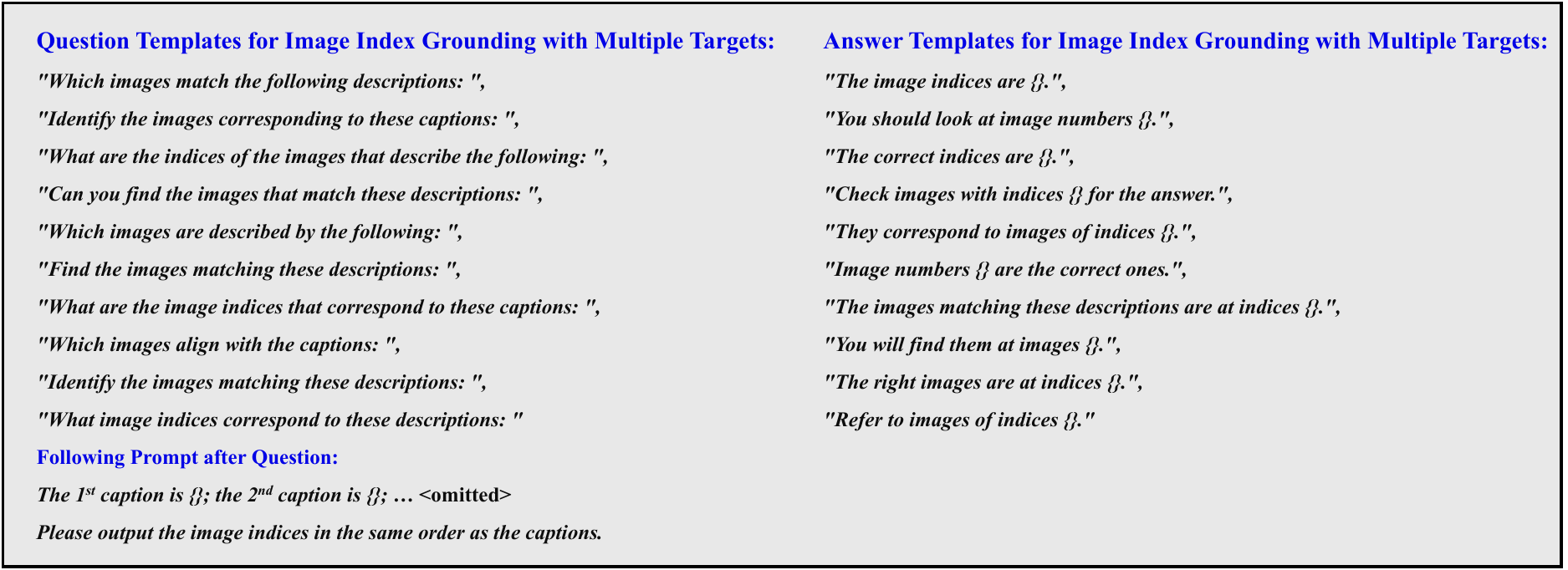}}
\caption{Instruction templates of image index grounding with multiple target images.}
\label{fig:supp_iig_multi}
% \vspace{-12pt}
\end{figure*}

\begin{figure*}[p]
\centering
% \vspace{-0.2cm}
\scalebox{1}{\includegraphics[width=1 \textwidth]{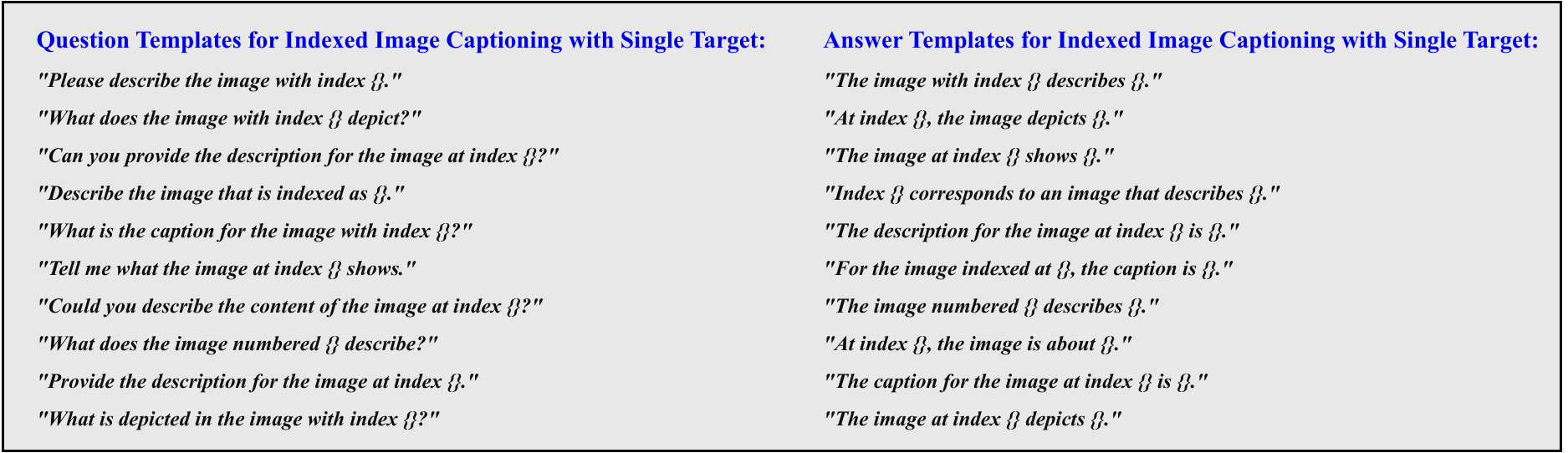}}
\caption{Instruction templates of indexed image captioning with single target image.}
\label{fig:supp_iic_single}
% \vspace{-12pt}
\end{figure*}

\begin{figure*}[ht]
\centering
% \vspace{-0.2cm}
\scalebox{1}{\includegraphics[width=1 \textwidth]{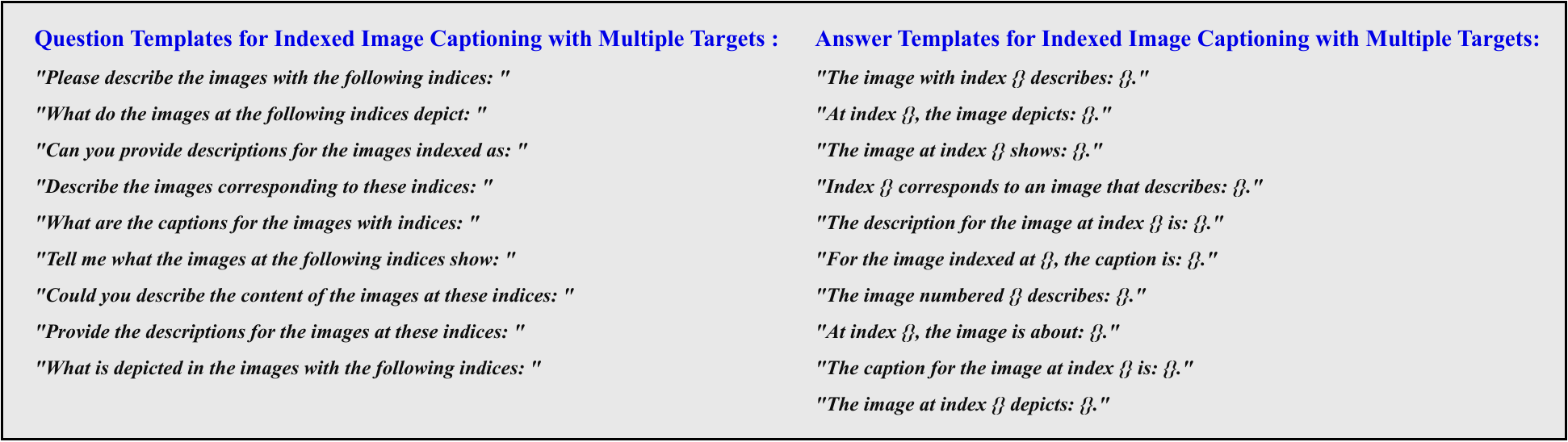}}
\caption{Instruction templates of indexed image captioning with multiple target images.}
\label{fig:supp_iic_multi}
% \vspace{-12pt}
\end{figure*}

\begin{figure*}[ht]
\centering
% \vspace{-0.2cm}
\scalebox{1}{\includegraphics[width=1 \textwidth]{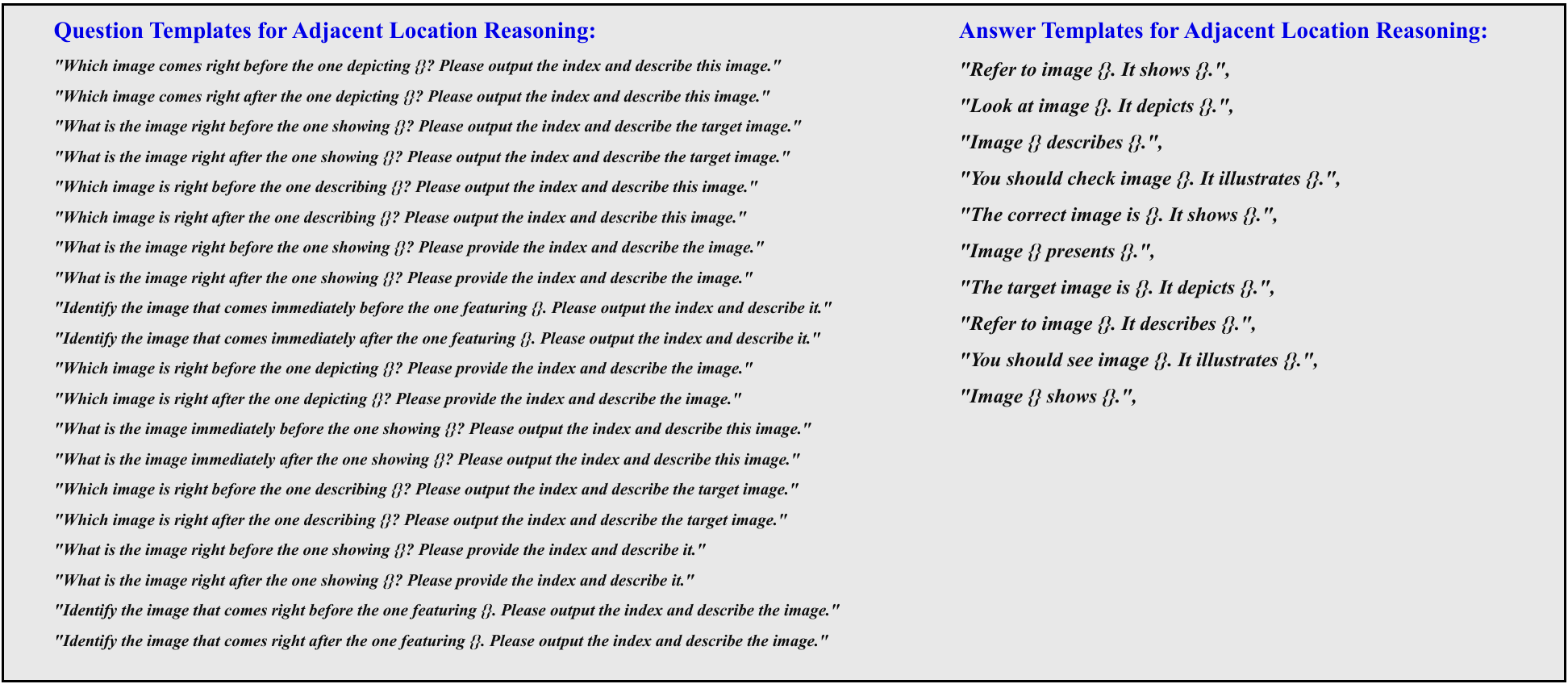}}
\caption{Instruction templates of adjacent location reasoning.}
\label{fig:supp_alr}
% \vspace{-12pt}
\end{figure*}

\noindent\textbf{Instruction Template Design.} To maximize the diversity and effectiveness of the training data, we design multiple instruction templates for each pretext task using GPT-4o~\cite{openai2023gpt4}. These templates include variations in both questions and answers to ensure robustness in model learning. Figures~\ref{fig:supp_iig_single}, \ref{fig:supp_iig_multi}, \ref{fig:supp_iic_single}, \ref{fig:supp_iic_multi}, and~\ref{fig:supp_alr} illustrate examples of these templates for different tasks, such as Image Index Grounding (IIG), Indexed Image Captioning (IIC), and Adjacent Location Reasoning (ALR). This variety enhances the model's ability to generalize across different types of instructions and strengthens its visual-language understanding capabilities.

\subsection{Clip Sequence}
\noindent\textbf{Data Source.} We utilize the Kinetics-700 dataset~\cite{carreira2019k700}, a large-scale video understanding dataset comprising approximately 650,000 ten-second video clips annotated with 700 distinct human action classes. The dataset offers diverse and naturalistic videos capturing a wide range of everyday activities. For our \method, we focus on the training split, which contains about 550,000 clips, ensuring a broad spectrum of visual and temporal information for model training.

\begin{figure*}[ht]
\centering
% \vspace{-0.2cm}
\scalebox{1}{\includegraphics[width=1 \textwidth]{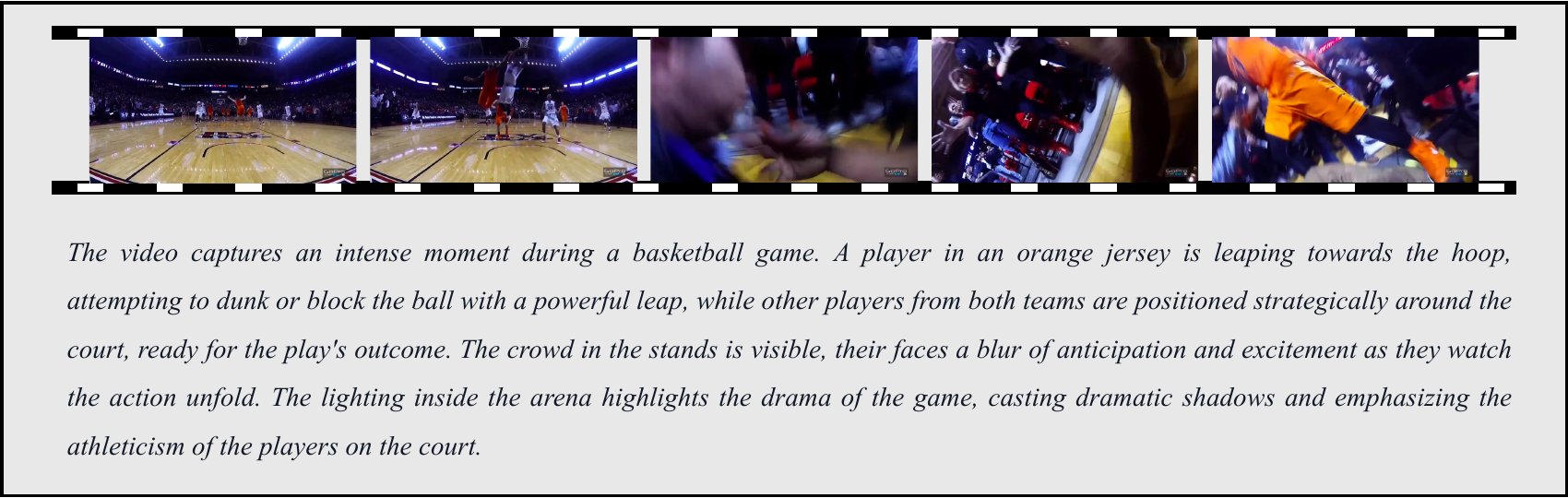}}
\caption{Instruction of an example in K700-LongVA-Captions.}
\label{fig:supp_good_case_longva}
% \vspace{-12pt}
\end{figure*}

\noindent\textbf{Clip Captioning with LongVA.} To generate textual descriptions for the video clips, we employ LongVA~\cite{zhang2024longva}, a state-of-the-art video captioning model. LongVA generates captions based on the visual content and the associated action label, as demonstrated in Figure~\ref{fig:supp_good_case_longva}. For instance, it accurately captures detailed descriptions like ``\texttt{A player in an orange jersey is leaping towards the hoop...}'' for basketball action clips. These captions provide contextual insights into the actions and environments depicted in the videos, enriching the clip sequence data with semantic annotations.

\begin{figure*}[ht]
\centering
% \vspace{-0.2cm}
\scalebox{1}{\includegraphics[width=1 \textwidth]{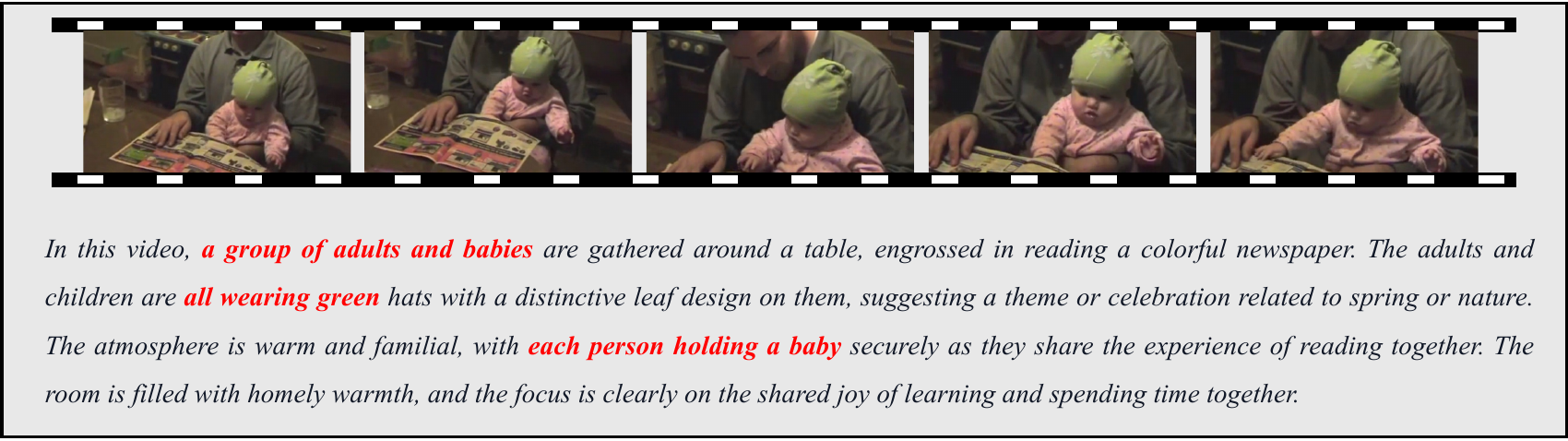}}
\caption{Instruction of a low-quality example in K700-LongVA-Captions. The incorrect descriptions are indicated by \textcolor{red}{\textbf{bold red }}text.}
\label{fig:supp_failure_case_longva}
% \vspace{-12pt}
\end{figure*}

Despite its strengths, LongVA occasionally produces hallucinated descriptions, as illustrated in Figure~\ref{fig:supp_failure_case_longva}. For example, a video showing an adult and a baby reading by a table might be misinterpreted as ``\texttt{a group of adults and babies...}'' Such hallucinations can potentially introduce noise into the dataset, affecting the reliability of the generated sequences. This highlights a trade-off between the scalability of automatic captioning and the quality of annotations.

% The combination of image and clip sequences provides complementary benefits. Image sequences offer precise visual-language alignment through detailed captions, enhancing the model's ability to link static visual elements to textual descriptions. Clip sequences, on the other hand, align more closely with real-world scenarios by capturing temporal dynamics in video data. Together, these datasets allow our \method to learn both fine-grained and temporally extended relationships, crucial for tasks such as temporal grounding and dense video captioning.

\section{Visualization}
\label{supp: vis}
In this section, we present qualitative results to demonstrate the effectiveness of our \method. Compared with the baseline model TimeChat, \method significantly improves upon repetitive text patterns, temporal localization accuracy, and the precision of event descriptions. 

In Figure~\ref{fig:supp_vis_ex1}, we analyze a video illustrating the instructions for making pancakes. TimeChat struggles to recognize ingredients in certain steps, providing incomplete descriptions such as ``\texttt{a bowl of eggs}'' and ``\texttt{a bowl of flour,}'' and its temporal predictions are imprecise, resulting in a fragmented understanding of the cooking process. In contrast, \method generates accurate and comprehensive event descriptions with precise temporal annotations, such as predicting ``\texttt{The next step is to mix the ingredients together in a bowl}'' as occurring between 25.0 to 61.4 seconds, closely aligning with the ground truth annotation: ``\texttt{mix flour, sugar, baking powder, and salt together in a bowl}'' from 25 to 61 seconds. This comparison highlights \method's ability to capture finer-grained details, maintain alignment with the video's temporal structure, and robustly capture the sequential flow of actions, effectively reducing redundancies and hallucinations often present in TimeChat's outputs. These improvements underscore \method's effectiveness in enhancing temporal grounding and understanding for long videos.

In Figure~\ref{fig:supp_vis_ex2}, a video illustrating the steps for cooking eggs is analyzed. TimeChat exhibits several issues, including generating repetitive text such as ``\texttt{She pours some into the bowl and stirs it}'' across multiple timestamps and misrecognizing objects, as seen in the caption ``She sets the bowl on the stove,'' where the ``\texttt{bowl}'' should be the ``\texttt{pan.}'' In contrast, \method minimizes these repetitive patterns and provides a more coherent description of the cooking process. Although \method incorrectly predicts the timestamp for the event ``\texttt{She stirs the mixture again,}'' it successfully captures the overall sequence of actions with detailed and contextually appropriate descriptions. This example highlights \method's enhanced capability to handle complex sequences with improved temporal and semantic accuracy compared to TimeChat.

In Figure~\ref{fig:supp_vis_ex3}, the video illustrates the steps for making tacos. TimeChat fails to recognize any of the correct cooking steps, producing entirely incorrect predictions and hallucinating that the video involves baking in an oven. In contrast, our \method provides general but accurate descriptions of the steps, despite omitting some specific ingredient details. For example, \method predicts ``\texttt{The man adds more ingredients to the dish and continues to stir it together}'' between 148.1 and 259.2 seconds. While this aligns with the overall sequence, the ground truth includes finer-grained annotations, such as adding specific ingredients like cumin powder and beef.

Figure~\ref{fig:supp_vis_ex4} presents a video of two girls preparing tofu soup. Our \method successfully captures key cooking steps, such as describing ``\texttt{They mix the ingredients in a pot and stir it}'' from 136.1 to 211.4 seconds. This aligns closely with the ground truth annotation, ``\texttt{add the tofu chunks and dissolve miso paste in the soup,}'' which spans 142 to 184 seconds. In contrast, TimeChat only captures the initial scene of the video, which is less relevant to the cooking process, failing to identify any meaningful steps.

Figure~\ref{fig:supp_vis_ex5} shows a video depicting a woman making cakes. TimeChat fails to generate relevant captions, producing incorrect descriptions, such as ``\texttt{a young man baking something.}'' While \method accurately identifies basic events within the video, providing correct and relevant captions that align with the overall context.

Finally, Figure~\ref{fig:supp_vis_ex6} shows a video of a woman preparing a salad in the kitchen. Both TimeChat and \method manage to produce correct captions for this video. However, neither model provides detailed step-by-step instructions. Instead, they capture only general actions, such as ``\texttt{woman talks}'' or ``\texttt{show the different ingredients,}'' overlooking specific details of the cooking steps. This indicates an area where further refinement of both models could enhance their performance in capturing detailed procedural actions.

These examples collectively demonstrate the superiority of our \method in reducing hallucinations, improving temporal alignment, and providing more accurate descriptions compared to TimeChat. However, they also highlight opportunities for improvement, such as better capturing fine-grained details and enhancing object recognition in complex cooking scenarios.

\section{Limitation Analysis} 
\begin{figure*}[ht]
\centering
% \vspace{-0.2cm}
\scalebox{1}{\includegraphics[width=1 \textwidth]{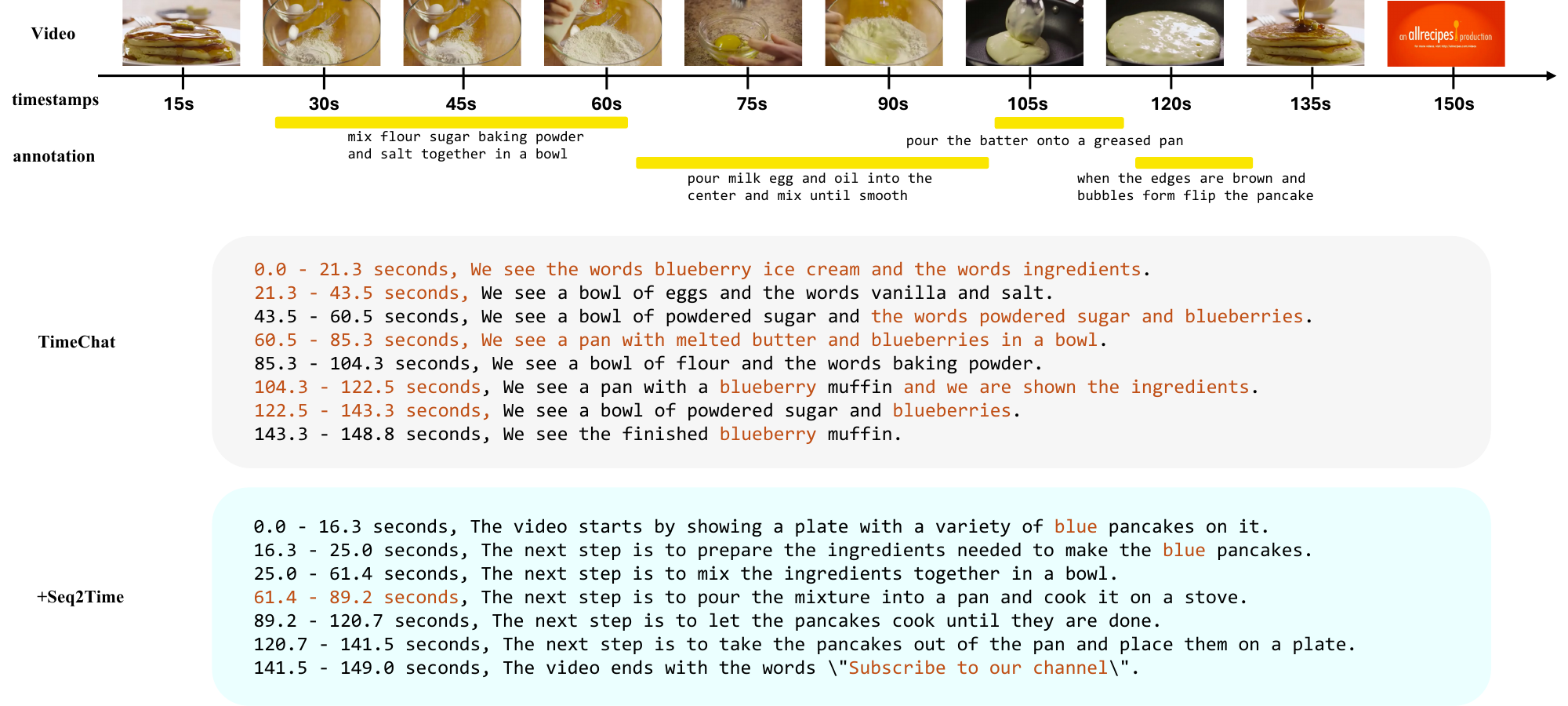}}
\caption{Qualitative example of our \method on TimeChat. The incorrect descriptions and timestamps are indicated by \textcolor{brown}{brown} text.}
\label{fig:supp_vis_ex1}
% \vspace{-12pt}
\end{figure*}

\begin{figure*}[ht]
\centering
% \vspace{-0.2cm}
\scalebox{1}{\includegraphics[width=1 \textwidth]{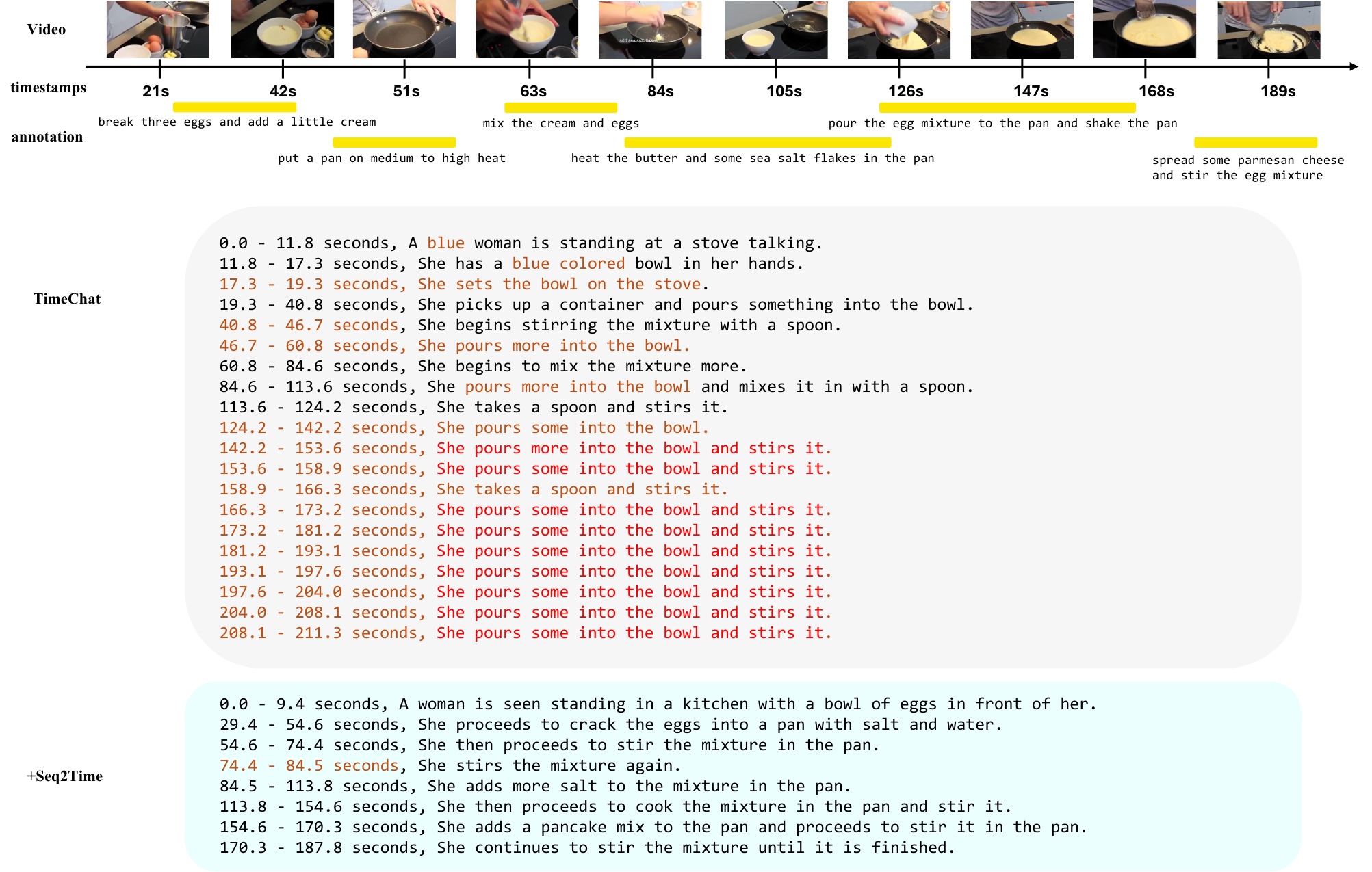}}
\caption{Qualitative example of our \method on TimeChat. The incorrect descriptions and timestamps are indicated by \textcolor{brown}{brown} text and repetitive text is highlighted in \textcolor{red}{red}.}
\label{fig:supp_vis_ex2}
% \vspace{-12pt}
\end{figure*}

\begin{figure*}[ht]
\centering
% \vspace{-0.2cm}
\scalebox{1}{\includegraphics[width=1 \textwidth]{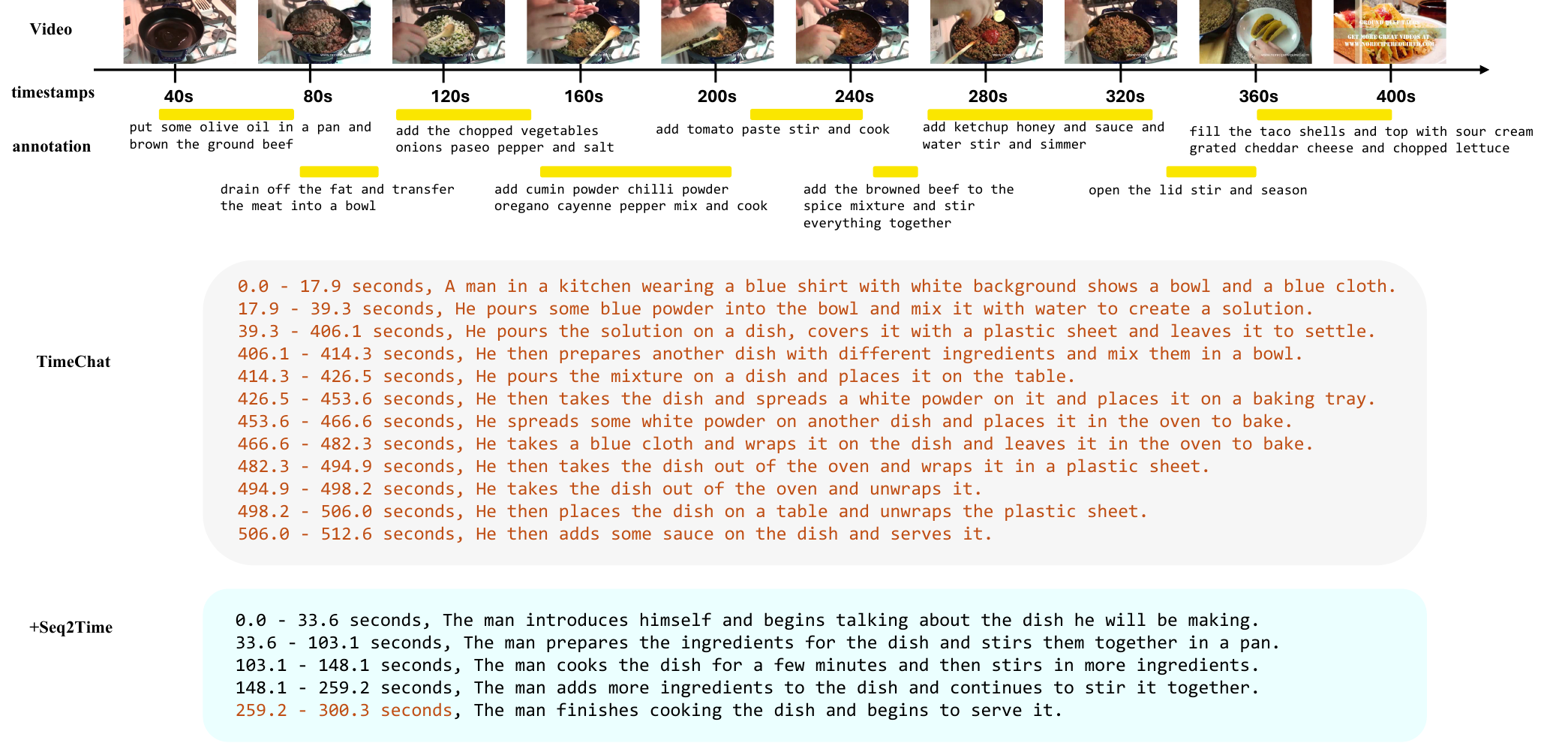}}
\caption{Qualitative example of our \method on TimeChat. The incorrect descriptions and timestamps are indicated by \textcolor{brown}{brown} text.}
\label{fig:supp_vis_ex3}
% \vspace{-12pt}
\end{figure*}

\begin{figure*}[ht]
\centering
% \vspace{-0.2cm}
\scalebox{1}{\includegraphics[width=1 \textwidth]{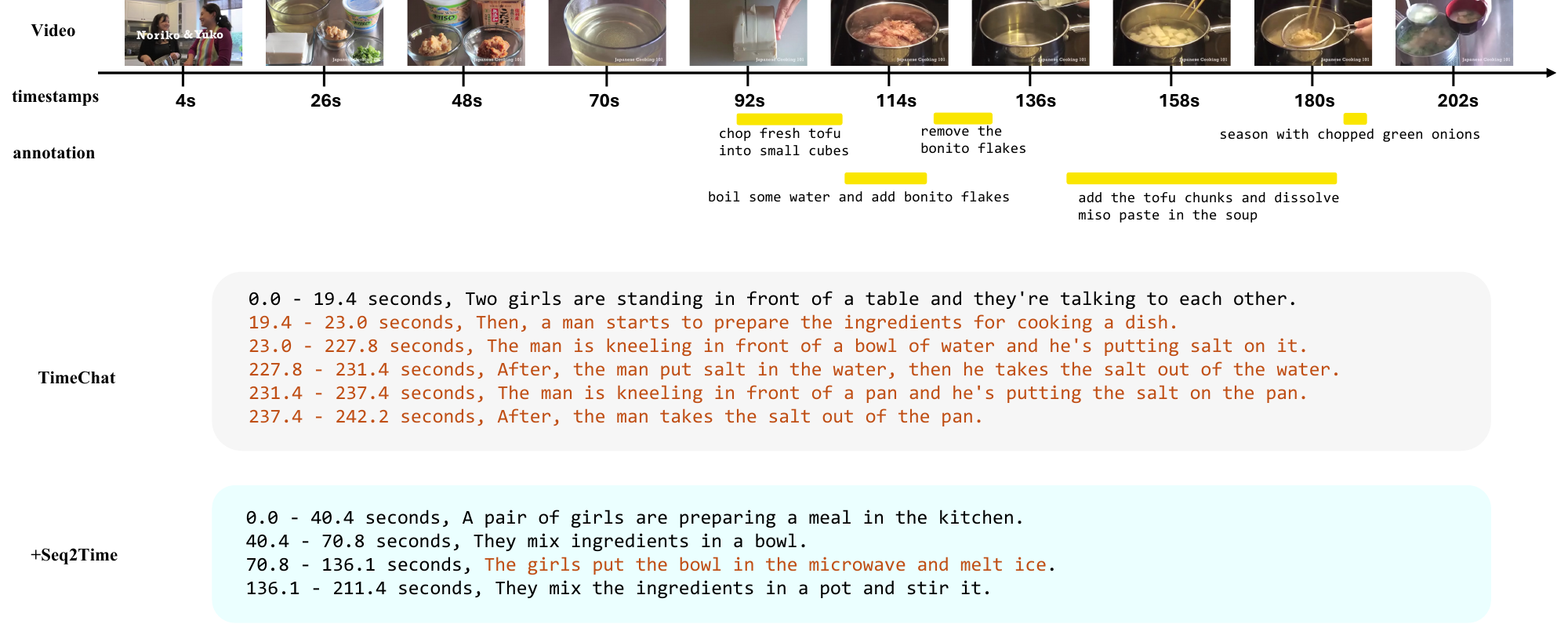}}
\caption{Qualitative example of our \method on TimeChat. The incorrect descriptions and timestamps are indicated by \textcolor{brown}{brown} text.}
\label{fig:supp_vis_ex4}
% \vspace{-12pt}
\end{figure*}

\begin{figure*}[ht]
\centering
% \vspace{-0.2cm}
\scalebox{1}{\includegraphics[width=1 \textwidth]{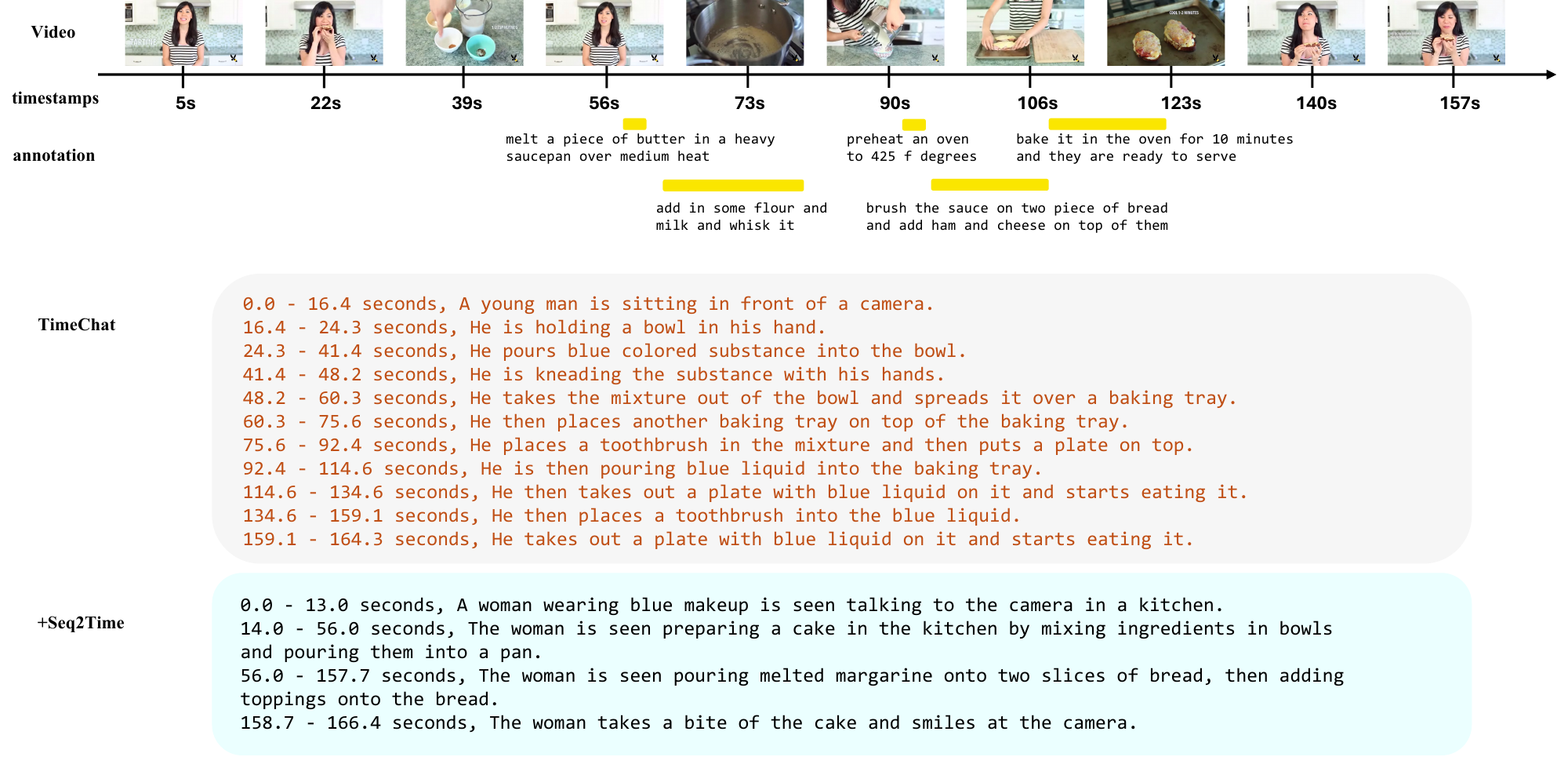}}
\caption{Qualitative example of our \method on TimeChat. The incorrect descriptions and timestamps are indicated by \textcolor{brown}{brown} text.}
\label{fig:supp_vis_ex5}
% \vspace{-12pt}
\end{figure*}

\begin{figure*}[ht]
\centering
% \vspace{-0.2cm}
\scalebox{1}{\includegraphics[width=1 \textwidth]{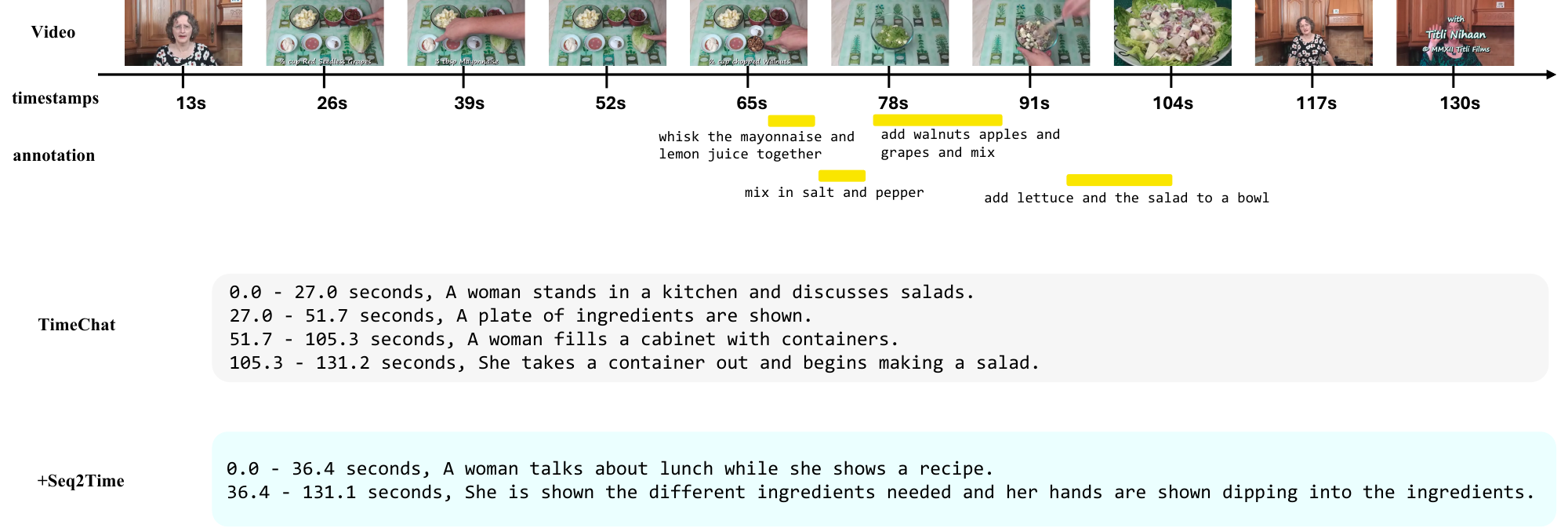}}
\caption{Qualitative example of our \method on TimeChat. The incorrect descriptions and timestamps are indicated by \textcolor{brown}{brown} text.}
\label{fig:supp_vis_ex6}
% \vspace{-12pt}
\end{figure*}

From Figure~\ref{fig:supp_vis_ex1} to Figure~\ref{fig:supp_vis_ex6}, we observe the overall effectiveness of our \method in capturing key events and providing accurate temporal annotations. However, some limitations remain, which could be addressed in future work. 

First, while the model generates correct captions with precise timestamps, the described events are occasionally less critical within the context of the video. For example, in Figure~\ref{fig:supp_vis_ex4}, ``\texttt{0.0 - 40.4 seconds, A pair of girls are preparing a meal in the kitchen,}'' and in Figure~\ref{fig:supp_vis_ex5}, ``\texttt{158.7 - 166.4 seconds, The woman takes a bite of the cake and smiles at the camera,}'' both describe valid events, but these are not essential steps in the respective cooking procedures. 

Second, the model sometimes overlooks fine-grained steps that are crucial for understanding detailed processes. For instance, in Figure~\ref{fig:supp_vis_ex3}, the model predicts a general event: ``\texttt{33.6 - 103.1 seconds, The man prepares the ingredients for the dish and stirs them together in a pan.}'' While this is accurate at a high level, it misses finer details such as ``\texttt{put olive oil,}'' ''\texttt{drain off the fat,}'' and ``\texttt{add chopped vegetables,}'' which are key preparatory steps in making tacos. Addressing these limitations would enhance the granularity and relevance of the captions generated by \method.

\clearpage

\clearpage

\clearpage

\clearpage

\clearpage

\end{document}